%% file: camera-ready.tex
\documentclass[letterpaper]{article} 
\usepackage{aaai2026}  
\usepackage{times}  
\usepackage{helvet}  
\usepackage{courier}  
\usepackage[hyphens]{url}  
\usepackage{graphicx} 
\usepackage{natbib}  
\usepackage{caption} 
\usepackage[linesnumbered,ruled,vlined]{algorithm2e}
\definecolor{AlgBlue}{RGB}{0,85,164}
\definecolor{AlgRed}{RGB}{239,65,53}
\definecolor{AlgGreen}{RGB}{100,161,56}
\definecolor{ForestGreen}{rgb}{0.13, 0.55, 0.13}

\SetKwComment{Comment}{\color{AlgGreen}\% }{}

\newcommand{\assign}{\leftarrow}

\newcommand{\FuncCall}[2]{\texttt{\bfseries #1(#2)}}
\SetKwProg{Function}{function}{}{}

\usepackage[table]{xcolor}
\usepackage{nicematrix}
\usepackage{acronym}
\usepackage{lineno}
\usepackage{booktabs}
\usepackage{comment}
\usepackage{multirow}
\usepackage{threeparttable}
\usepackage[usestackEOL]{stackengine}
\usepackage{makecell}
\usepackage{enumitem}
\usepackage{multicol}

\newcommand{\ie}{{\it i.e.}}

\newcommand{\ML}{machine learning}
\newcommand{\otoprule}{\midrule[\heavyrulewidth]}
\definecolor{BlueGray2}{rgb}{ .918,  .925,  .945}
\newcommand{\ul}{\underline}
\newcommand{\parab}[1]{{\bf #1}}

\definecolor{Gray}{HTML}{EAECF1}
\definecolor{Brown}{HTML}{964B00}
\definecolor{Green}{HTML}{70BE71}
\definecolor{White}{RGB}{255,255,255}
\setlist[description]{leftmargin=\parindent,labelindent=\parindent}
\newcommand{\pattern}{\textsc{Pattern}\xspace}

\title{Poisoning with A Pill: Circumventing Detection in Federated Learning}
\author{
    Hanxi Guo\textsuperscript{\rm 1}\thanks{Work done while Hanxi Guo was a master’s student at SJTU, prior to joining Purdue University.}, 
    Hao Wang\textsuperscript{\rm 2}, 
    Tao Song\textsuperscript{\rm 3}\thanks{Tao Song is the corresponding author.}, 
    Tianhang Zheng\textsuperscript{\rm 4}, \\
    Yang Hua\textsuperscript{\rm 5},
    Haibing Guan\textsuperscript{\rm 3},
    Xiangyu Zhang\textsuperscript{\rm 1}
}
\affiliations{
    \textsuperscript{\rm 1}Purdue University\\
    \textsuperscript{\rm 2}Stevens Institute of Technology\\
    \textsuperscript{\rm 3}Shanghai Jiao Tong University \\
    \textsuperscript{\rm 4}Zhejiang University \\
    \textsuperscript{\rm 5}Queen’s University Belfast \\
    guo778@purdue.edu, hwang9@stevens.edu, songt333@sjtu.edu.cn, zthzheng@zju.edu.cn, \\ Y.Hua@qub.ac.uk, hbguan@sjtu.edu.cn, xyzhang@purdue.edu
}

\begin{document}

\maketitle

\begin{abstract}
Federated learning (FL) protects data privacy by enabling distributed model training without direct access to client data. However, its distributed nature makes it vulnerable to model and data poisoning attacks. While numerous defenses filter malicious clients using statistical metrics, they overlook the role of model redundancy, where not all parameters contribute equally to the model and attack performance. Current attacks manipulate all model parameters uniformly, making them more detectable, while defenses focus on the overall statistics of client updates, leaving gaps for more sophisticated attacks. We propose an attack-agnostic augmentation method to enhance the stealthiness and effectiveness of existing poisoning attacks in FL, exposing flaws in current defenses and highlighting the need for fine-grained FL security. Our three-stage methodology, including \textit{pill construction}, \textit{pill poisoning}, and \textit{pill injection}, injects poison into a compact subnet (\ie, pill) of the global model during the iterative FL training. Experimental results show that FL poisoning attacks enhanced by our method can bypass 8 state-of-the-art (SOTA) defenses, gaining an up to 7x error rate increase, as well as on average a more than 2x error rate increase on both IID and non-IID data, in both cross-silo and cross-device FL systems.
\end{abstract}

\begin{links}
    \link{Code}{https://github.com/MarkGHX/PoisonPill}
\end{links}

\section{Introduction}
\label{sec:intro}
With the rising demand for machine learning and cloud computing, Federated Learning (FL)\cite{konevcny2016federated,mcmahan2017communication} has emerged as a key approach for training models on distributed data from scattered clients. Unlike centralized \ML, FL avoids direct data access, reducing communication overhead and enhancing privacy. However, its distributed nature leaves it vulnerable when clients are compromised. Numerous studies~\cite{baruch2019little,fang2020local,bhagoji2019analyzing,shejwalkar2021manipulating,cao2022mpaf,bagdasaryan2020backdoor} have examined \textit{poisoning attacks}, where malicious clients manipulate the global model. These fall into two categories: 1) \textit{Model poisoning} which directly alters local updates to skew global parameters~\cite{fang2020local,shejwalkar2021manipulating}; and 2) \textit{Data poisoning} that injects malicious samples into local datasets~\cite{bagdasaryan2020backdoor,tolpegin2020data,xie2020dba,sun2019can,wang2020attack,chen2017targeted,liu2018trojaning,qi2022sra}. Such attacks threaten FL’s integrity and reliability~\cite{lyu2020threats,kairouz2021advances}.

To mitigate these attacks, defenses have been proposed, including \textit{adaptive client filtering}~\cite{krum,cao2020fltrust,xu2021signguard,nguyen2022flame,yan2023skymask}, \textit{statistical parameter aggregation}~\cite{yin2018trim,bulyan,foolsgold,panda2022sparsefed,han2023towards}, \textit{client-dominant detection}~\cite{guo2021siren,guo2024siren+,sun2021flwbc,zhang2023flip,zhu2023leadfl,park2023feddefender}, and \textit{other advanced metrics and pipelines}~\cite{xie2019zeno,xie2021crfl,cao2023fedrecover,cao2022flcert,zhang2022fldetector}. 
These approaches aim to identify suspicious updates, which are usually evident in model poisoning attacks that uniformly alter parameters.

We argue that modifying all parameters uniformly is not a cost-effective approach. Studies on model pruning~\cite{frankle2018lottery,lin2018deep,han2015deep,mugunthan2022fedltn,jiang2022model} show that parameters do not contribute equally to a model's performance. Altering \textit{redundant} parameters wastes resources and reduces attack \textit{stealthiness}. A more effective strategy is to target \textit{critical} parameters~\cite{zhang2023oblivion}, which significantly impact performance, thereby increasing the attack's effectiveness while maintaining stealthiness. Thus, we propose a novel attack-agnostic augmentation method that enhances model poisoning attacks using a three-stage pipeline: {\em pill construction}, {\em pill poisoning}, and {\em pill injection}. In the first stage, we design a pill blueprint and identify its corresponding subnet instance in the target model. During {\em pill poisoning}, existing FL attacks are applied in an attack-agnostic manner to poison the selected pill. Finally, in {\em pill injection}, the poisoned pill is inserted into an estimated benign update, and a two-step adjustment is used to minimize the difference between the poisoned and benign updates. This approach dynamically generates, poisons, and injects a pill into the global model, augmenting existing FL poisoning attacks.

We conduct extensive experiments to evaluate the effectiveness of our augmentation method. We apply it to four baseline poisoning attacks: {sign-flipping attack}, {trim attack}~\cite{fang2020local}, {krum attack}~\cite{fang2020local}, and {min-max attack}~\cite{shejwalkar2021manipulating}. Using both the original and augmented versions, we measure error rates (\ie, the proportion of incorrect predictions) of the global model trained with nine aggregation rules: FedAvg~\cite{mcmahan2017communication}, FLTrust~\cite{cao2020fltrust}, Multi-Krum~\cite{krum}, Median~\cite{yin2018trim}, Trim~\cite{yin2018trim}, Bulyan~\cite{bulyan}, FLDetector~\cite{zhang2022fldetector}, DnC~\cite{shejwalkar2021manipulating}, and Flame~\cite{nguyen2022flame}. These aggregation rules represent most existing defense metrics. We also design an adaptive defense where the defender has full knowledge of our pipeline and implementation. Results show our method substantially improves existing FL poisoning attacks, leading to over a 2x average increase in model prediction error rates under existing defenses, and up to a 7x increase in some cases.

Our contributions are summarized as follows:
\begin{itemize}
    \item We propose a generic, attack-agnostic augmentation method that enhances poisoning attacks against robust FL by encapsulating model poisoning attacks into well-defined subnets (\ie, pills) with comprehensive metric-based adjustments.
    \item Extensive experiments on three common datasets against nine aggregation rules demonstrate that our method helps baseline attacks bypass almost all existing defenses, which cannot be attacked by original versions.
    \item We identify limitations of existing poisoning attacks and defenses in FL, highlighting the need and potential for fine-grained FL security.
\end{itemize}

\section{Background and Related Work}
\label{sec:background}
\subsection{Federated Learning}
Federated Learning (FL)~\citep{konevcny2016federated,mcmahan2017communication} trains a global model using the information from a swarm of clients without the direct access to each client's data. 
In a standard FL training process, within an arbitrary communication round $t$, the FL server first distributes its global model $\boldsymbol{g}_t$ to all the clients $K$. 
After receiving this global model, each client $i$ trains a local model $\boldsymbol{g}_t^{(i)}$ with its local data $D^{(i)}$, and uploads the model update $\Delta\boldsymbol{g}_t^{(i)}$ to the FL server.
After receiving the model updates from the clients, the FL server uses aggregation rules to calculate the global model $\boldsymbol{g}_{t+1}$ for the next round.
The objective of FL can be formulated as:
\begin{equation}
    \mathop{min}_{\boldsymbol{g}} \sum_{i=0}^K \frac{|D^{(i)}|}{|D|}\cdot f(D^{(i)}, \boldsymbol{g}).
\end{equation}

\subsection{Poisoning Attacks in FL}
Following prior studies~\citep{shejwalkar2022back, khan2023pitfalls, jere2020taxonomy}, poisoning attacks in Federated Learning (FL) fall into two categories: {\em model poisoning} and {\em data poisoning}. In {\em model poisoning attacks}, adversaries compromise the global model by directly altering local model updates~\citep{baruch2019little,fang2020local,shejwalkar2021manipulating,cao2022mpaf,bhagoji2019analyzing}. In {\em data poisoning attacks}, they corrupt local datasets to indirectly affect the global model~\citep{tolpegin2020data,bagdasaryan2020backdoor,xie2020dba,sun2019can,wang2020attack,zhang2022neurotoxin}. Further details are in the extended version.

Our pill design draws inspiration from the {\em subnet replacement attack} (SRA)~\citep{qi2022sra}, a backdoor injection method that limits backdoors to a small subnetwork. SRA trains this subnet with tainted data, substitutes the target model's parameters, and disconnects the subnet to maintain attack effect. Drawing from SRA’s stealthy yet potent design, we propose a heterogeneous-width {\em pill blueprint} for varied FL poisoning attacks. Unlike SRA’s one-time injection, our approach gradually poisons the global model during training, improving robustness against various defenses.

\subsection{Defenses against Poisoning Attacks in FL}
Existing defenses can be categorized based on the mitigation strategies that they utilize, including \textit{Adaptive Client Filtering}, \textit{Statistical Parameter Aggregation}, \textit{Client-dominant Detection}, and \textit{Other Advanced Metrics and Pipelines}. To comprhensively evaluate our method, we use \textit{Multi-Krum (MKrum)}~\citep{krum}, \textit{Trimmed Mean (Trim)}~\citep{yin2018trim}, \textit{Coordinate-wise Median (Median)}~\citep{yin2018trim}, \textit{Bulyan}~\citep{bulyan}, \textit{FLTrust}~\citep{cao2020fltrust}, \textit{FLDetector (FLD)}~\citep{zhang2022fldetector}, \textit{DnC}~\citep{shejwalkar2021manipulating}, and \textit{Flame}~\citep{nguyen2022flame}, a set of representative defenses, as our baselines. More details are in the extended version.

\subsection{Threat Model}
\label{sec:threat_model}
We follow the typical threat model used in existing studies~\citep{fang2020local, shejwalkar2021manipulating}, where the attacker has access to a subset of compromised clients and aims to increase the error rates of the global model on specific classes or across all classes. In this scenario, defenses cannot directly analyze the data on each client as the defender's setting in ~\cite{krum,yin2018trim,cao2020fltrust,guo2021siren}. Instead, they identify malicious clients by analyzing the uploaded client updates. Further details are in the extended version.

\begin{figure*}[t]
    \centering
    \includegraphics[width=0.85\textwidth]{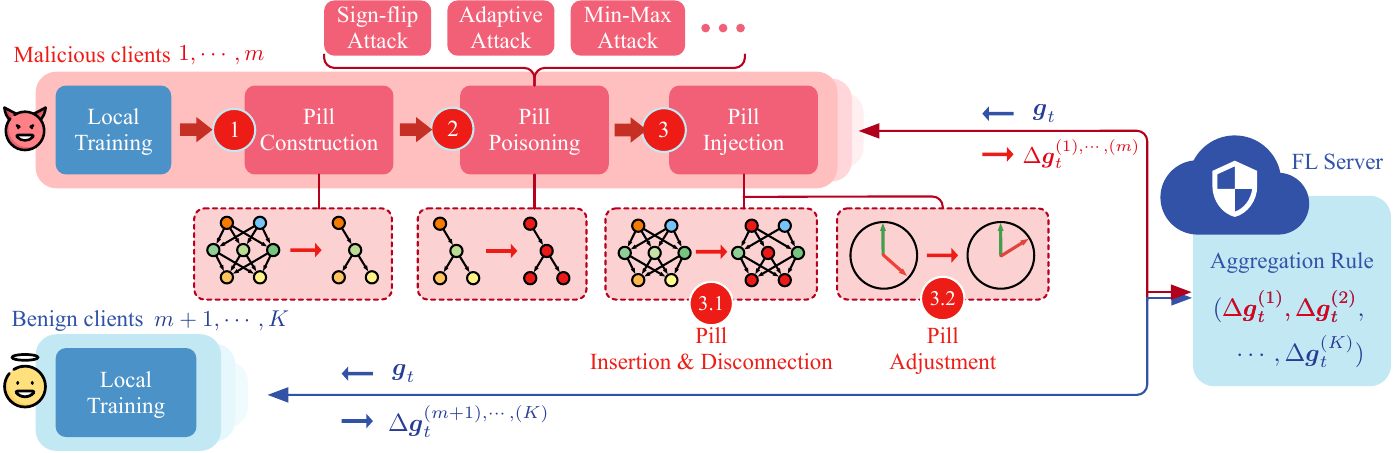}
    \caption{Overview of our augmentation method. The {\color{red} red} parts indicate our augmentation method's contribution, and the {\color{cyan} cyan} parts represent the standard federated learning architecture.}
    \label{fig:overview}
\end{figure*}
\section{Design Objectives and Challenges}
\label{sec:challenge}
After analyzing the drawbacks and various implementations of existing FL poisoning attacks, we define three main objectives for our attack augmentation method:
1) For \textit{stealthiness}, the augmentation method should stay stealthy 
while achieving 
comparable performance with original attacks. 
2) For \textit{compatibility}, the augmentation should be compatible with most of the existing FL poisoning attacks with few modifications on their implementations. 
3) For \textit{generality}, the attack augmentation should be able to bypass general detection methods with different detection metrics. 

Corresponding to each objective, three challenges need to be addressed:

\begin{itemize}
    \item It presents a significant challenge that the 
    attack augmentation method must use significantly fewer parameters while still achieving similar attack effects. 
    \item It is 
    challenging to develop a uniform augmentation method for 
    various FL poisoning attacks since they require different information and are implemented in different training stages. 
    \item It is difficult
    to devise a general strategy that bypasses all common
    detection approaches, while  guaranteeing the attack effectiveness.
\end{itemize}

\section{Design}
\label{sec:design}

\subsection{Overview of Our Method}
We are the first that propose a universal attack augmentation pipeline for most FL poisoning attacks, considering all of the \textit{stealthiness}, \textit{compatibility}, and \textit{generality}. Figure~\ref{fig:overview} presents the three key stages. 
In the extended version, we include a comprehensive table of all main symbol notations used in the paper, along with an algorithm table detailing the full workflow of our method. 

\textbf{Stage {\large\textcircled{\small 1}}}: \textbf{Pill Construction}. It leverages a dynamic subnetwork search algorithm to achieve \textit{stealthiness} by selecting the tiny pill from the global model $\boldsymbol{g}_t$ based on the importance of model parameters. Additionally, six dynamic search patterns are designed to prevent being traced.

\textbf{Stage {\large\textcircled{\small 2}}}: \textbf{Pill Poisoning}. In this state, we reapply existing FL poisoning attacks to the selected poison pill, using an extra trained model $\boldsymbol{\hat{g}^{m}_{t+1}}$ (trained on data from the compromised clients) as the attacker's base model. For \textit{compatibility}, we only modify the input of the existing FL poisoning attacks and utilize their outputs, without any interference to their internal implementations. This black-box utilization lets our method be attack-agnostic and compatible with most of the existing FL poisoning attacks.

\textbf{Stage {\large\textcircled{\small 3}}}: \textbf{Poison Pill Injection}. It contains pill insertion \& disconnection, and pill adjustment. In this stage, our augmentation method injects the poison pill into the estimated benign update $\Delta\boldsymbol{\widetilde{g}}_{t+1}$, and further adjusts the boosting magnitude of both the poison pill parameters and the remaining parameters. We propose a two-step dynamic adjustment to enhance the \textit{generality} of our method against most defenses. 

\subsection{Pill Construction}
\label{subsec:select}
This stage aims to construct a pill structure for augmenting the \textit{stealthiness} while retaining the original attack effectiveness. The pill is carefully crafted to only involve a minimal subset of parameters from specific positions of the target model. 
We first define a pill's blueprint as the pill's graphic structure, independent of target model parameters.  
Then, we propose a dynamic pill search algorithm to identify and map concrete parameters from the target model to the blueprint. 

\parab{Designing Pill Blueprint.}
Inspired by SRA~\citep{qi2022sra}, which demonstrates that poisoning a narrow subnetwork (one neuron/channel per layer) can effectively implant backdoors in machine learning models (outside the FL setting), we adapt and generalize the concept for FL. The original SRA design is unsuitable for our goals as it employs a fixed, pre-selected subnet tailored to a specific architecture, ignores FL’s training dynamics, lacks adaptability to different targets, and produces non-stealthy subnets with disproportionately large weights to propagate the poison through a small network.
We introduce a novel blueprint method with a general subnet structure whose instantiations vary dynamically during FL training. A search algorithm selects important neurons at each step, enabling small weight modifications while ensuring effective dissemination and high stealthiness. The blueprint also supports multiple targets by simultaneously manipulating outputs across all relevant neurons.
Specifically, our pill blueprint design follows the rules below (more details shown in the extended version): 
\begin{enumerate}
    \item[1.] The pill blueprint only contains one neuron in each linear layer or one channel in each convolutional layer, except for the last two layers. Suppose $\mathcal{N}_i^p$ represents the neuron/channel number in Layer $i$ in our pill blueprint, then $\mathcal{N}_i^p = 1$ when $i<L-1$, where $L$ is the total layer counts in our pill blueprint.
    \item[2.] In the last two layers of our pill blueprint, $\mathcal{N}^p_{L-1} = \mathcal{N}^p_{L} =$ \textit{number of classes}.
\end{enumerate}

\begin{figure*}[t]
    \centering
    \begin{minipage}[t]{.45\linewidth}
        \centering
        \includegraphics[width=0.95\linewidth]{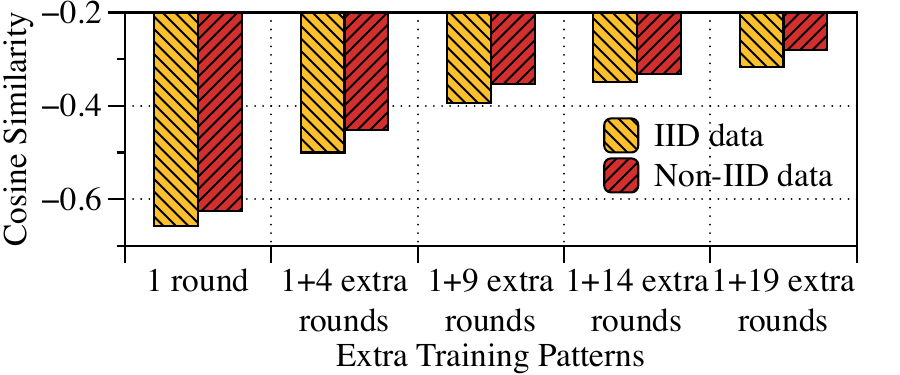}
        \caption{Cosine similarity between FLTrust server and malicious updates with different extra local epochs.}
        \label{fig:extra train}
    \end{minipage}
    \hfill
    \nextfloat
    \begin{minipage}[t]{.45\linewidth}
        \centering
        \includegraphics[width=0.78\linewidth]{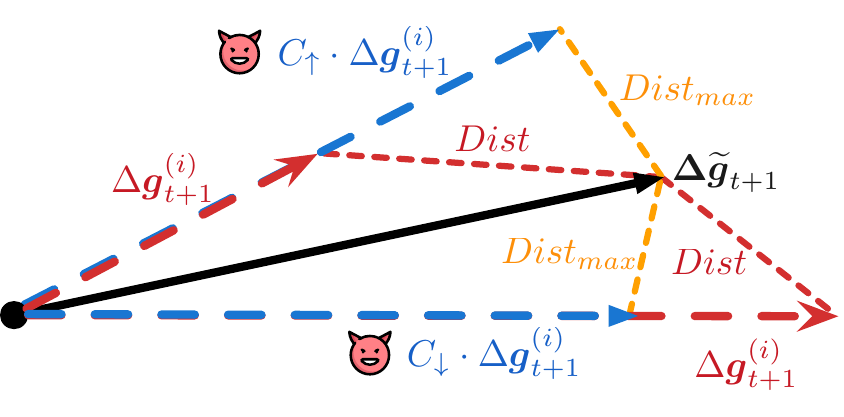}
        \caption{Intuition behind distance-based adjustment in our augmentation method.}
        \label{fig:dist adjust}
    \end{minipage}
\end{figure*}

\parab{Dynamic Pill Search.}
According to existing studies on neural network pruning~\citep{frankle2018lottery,lin2018deep,han2015deep,mugunthan2022fedltn,jiang2022model}, parameters with a larger magnitude typically dominate the model's performance. 
Thus, the best approach is to analyze the parameters to find a globally optimal pill that includes the most critical parameters.

However, such a globally optimal pill could be identified via a pruning-based method~\citep{wu2020mitigating, sun2021soteria}, and hence our attack could be easily detected. Besides, searching for a globally optimal pill is inefficient when the model has a large number of parameters. Thus, we search for an approximate pill instead, with an attacker-defined start point, and only evaluate a small subset of the entire model's parameters. We name the search algorithm as ``approximate max pill search''.
The key idea is to perform a targeted neuron search at each layer by focusing only on the neurons connected to the selected neurons from the previous layer, following a high-sum-of-weights-first principle that prioritizes neurons based on the cumulative sum of their connection weights to the previously selected neurons. The entire search contains four steps:

\textbf{Step 1. Random Start Point Selection:} Randomly select neurons from the first layer of the target model as start points, denoted as $\mathcal{V}_1$, based on the neuron count $\mathcal{N}^p_1$ in pill blueprint's first layer. Such selected start point neurons are fixed throughout the search.

\textbf{Step 2. Layer-wise Search:} For each subsequent layer $l_i$, we compute the sum of weights connecting neurons in $\mathcal{V}_{i-1}$ to neurons in $l_i$ and rank the neurons in $l_i$ based on the descending order of the sum of weights. Top $\mathcal{N}^p_{i}$ neurons are chosen for $\mathcal{V}_i$, where $\mathcal{N}^p_{i}$ denotes the number of neurons in the pill blueprint's $i$-th layer.

\textbf{Step 3. Output Neuron Pairing:} Pair the selected neurons $\mathcal{V}_{L-1}$ in the final hidden layer with the neurons in the output layer $l_L$, ensuring a one-to-one correspondence.

\textbf{Step 4. Pill Mask Construction:} Two masks are constructed. $M$ marks the pill parameters in the target model, and $M_{disc}$ records the disconnection locations between the pill and the remained neurons in the target model.

The searched pill ensures both effectiveness and stealthiness in attacks. Detailed information for each step is provided in the extended version, along with a concrete example and an overhead analysis of the pill search algorithm.

\subsection{Pill Poisoning}
In the \textbf{pill poisoning} stage, we aim to condense the poison into the pill using existing attacks.
To achieve compatibility, our method simply reuses existing FL poisoning attacks, without any intrusive modification to their original implementations. We only modify the input of existing FL poisoning attacks by replacing the base model update with the update from a model that has undergone extra training rounds, denoted as $\Delta\boldsymbol{\hat{g}^{m}_{t+1}}$. Additionally, we restrict changes to parameters within the pill.
The output is a poisoned pill 
that will be used in the next \textbf{pill injection} stage.

The motivation to use an extra-trained model update as the reference model update is shown in Figure~\ref{fig:extra train}. As shown in the figure, with the increasing number of extra training rounds on the malicious clients, the generated malicious model update becomes less opposite to the FLTrust~\citep{cao2020fltrust} 
server's model update. 
Thus, we adopt the extra training in our method and limit the extra training epoch number $E_{extra}$ to less than the number of malicious clients $m$ times the benign local training epoch number $E$, denoted as $E_{extra} \leq m \cdot E$. 
This constraint ensures compliance with the threat model, as the attacker can utilize the data and computational resources of compromised clients.

\subsection{Pill Injection}
\label{subsec:insert}
In the \textbf{pill injection} stage, we aim to inject the pill into the
model
and use a two-step adjustment method to further camouflage the pill. Thus, the entire injection stage could be divided into two parts - pill injection 
and camouflaging. 
After this stage, the poison pill is 
seamlessly integrated 
with 
the benign model update and uploaded to the FL server.

\begin{table*}[t]
	\centering
    \small
    \setlength{\tabcolsep}{1.8pt}
	    \begin{tabular}{cccccccc|ccccccc}
		\toprule
        \textbf{Data Distribution} & \multicolumn{7}{c}{\textbf{IID}} & \multicolumn{7}{c}{\textbf{Non-IID}}\\
        \otoprule
		\textbf{Attack} & \textbf{FedAvg} & \textbf{FLTrust} & \textbf{MKrum} & \textbf{Bulyan} & \textbf{Median} & \textbf{Trim} & \textbf{FLD} & \textbf{FedAvg} & \textbf{FLTrust} & \textbf{MKrum} & \textbf{Bulyan} & \textbf{Median} & \textbf{Trim} & \textbf{FLD}\\
		\otoprule
		\multirow{1}{*}{No Attack} & 0.109 & 0.107 & 0.105 & 0.105 & 0.123 & 0.106 & 0.115 & 0.113 & 0.115 & 0.115 & 0.112 & 0.142 & 0.115 & 0.122 \\
        \midrule
		\multirow{1}{*}{Sign-Flipping} & \textbf{0.943} & 0.114 & 0.108 & 0.126 & 0.136 & 0.116 & 0.118  & \textbf{0.917} & \textbf{0.126} & 0.117 & 0.132 & 0.152 & 0.124 & 0.127 \\
		\cmidrule{2-15}
        \rowcolor{BlueGray2}
        \cellcolor{White} + Poison Pill & 0.667 & \textbf{0.115} & \textbf{0.764} & \textbf{0.379} & \textbf{0.523} & \textbf{0.314} & \textbf{0.646} & 0.543 & 0.122 & \textbf{0.754} & \textbf{0.430} & \textbf{0.522} & \textbf{0.311} & \textbf{0.688} \\
		\midrule
		\multirow{1}{*}{Trim Attack} & 0.243 & 0.109 & 0.139 & 0.146 & 0.174 & 0.179 & \textbf{0.116} & 0.332 & 0.120 & 0.201 & 0.163 & 0.231 & \textbf{0.238} & 0.124\\
		\cmidrule{2-15}
        \rowcolor{BlueGray2}
        \cellcolor{White} + Poison Pill & \textbf{0.618} & \textbf{0.576} & \textbf{0.638} & \textbf{0.284} & \textbf{0.453} & \textbf{0.219} & 0.115 & \textbf{0.668} & \textbf{0.517} & \textbf{0.687} & \textbf{0.292} & \textbf{0.473} & 0.223 & \textbf{0.222} \\
		\midrule
		\multirow{1}{*}{Krum Attack} & 0.116 & 0.109 & 0.189 & 0.201 & 0.172 & 0.137 & \textbf{0.786} & 0.128 & 0.116 & 0.235 & 0.276 & 0.217 & 0.160 & \textbf{0.947}\\
		\cmidrule{2-15}
        \rowcolor{BlueGray2}
        \cellcolor{White} + Poison Pill & \textbf{0.735} & \textbf{0.155} & \textbf{0.715} & \textbf{0.422} & \textbf{0.578} & \textbf{0.310} & 0.637 & \textbf{0.716} & \textbf{0.151} & \textbf{0.737} & \textbf{0.468} & \textbf{0.730} & \textbf{0.334} & 0.690\\
        \midrule
        \multirow{1}{*}{Min-Max Attack} & 0.183 & 0.110 & 0.431 & \textbf{0.330} & 0.183 & 0.218 & \textbf{0.825} & 0.269 & 0.125 & \textbf{0.619} & \textbf{0.434} & 0.255 & 0.278 & \textbf{0.831}\\
		\cmidrule{2-15}
        \rowcolor{BlueGray2}
        \cellcolor{White} + Poison Pill & \textbf{0.702} & \textbf{0.303} & \textbf{0.668} & 0.327 & \textbf{0.514} & \textbf{0.314} & 0.778 & \textbf{0.629} & \textbf{0.320} & 0.612 & 0.406 & \textbf{0.547} & \textbf{0.376} & 0.822\\
		\bottomrule
	    \end{tabular}
    \caption{Error rates on Fashion-MNIST under cross-silo setting with 20\% malicious clients in the 50-client FL system.}
	\label{Table:fmnist-cs-20}
\end{table*}

\parab{Pill Insertion \& Disconnection.}
In this part, our goal is to insert the pill 
into the model, and minimize the impact of the benign model updates on our pill. We use an estimated global model update as the benign model update, which is estimated as the coordinate-wise mean values of all the normal model updates from the compromised clients. The estimation process is hence presented as Equations~(\ref{equation:estimation start}) (\FuncCall{Estimation}{} in Algorithm in the extended version),
\begin{align}
    \label{equation:estimation start}
     \Delta\boldsymbol{\widetilde{g}}_{t+1} &\assign \mathop{mean}\{\Delta\boldsymbol{g'}^{(1),\cdots,(m)}_{t+1}\},
\end{align}
where $\Delta\boldsymbol{g'}^{(i)}_{t+1}$ is the normal updates from the compromised clients. By aggregating information from multiple malicious clients, the estimated global model update is more similar to the genuine one, providing more budget for our poison pill.

After obtaining the estimated global model update $\boldsymbol{\Delta\widetilde{g}}_{t+1}$, we directly replace the 
parameters corresponding to the pill parameters (which have been poisoned in the previous stage) 
via the pill's mask $\boldsymbol{M}$. Then, we replace the parameters that connect the  pill and the other estimated global model updates with {\em the disconnection update} $\Delta\boldsymbol{g}_{t+1}^{zero}$, using the disconnection mask $\boldsymbol{M}_{disc}$. The disconnection update $\Delta\boldsymbol{g}_{t+1}^{zero}$ is calculated as $0 - \boldsymbol{g}_t$, and is bounded by the maximum and minimum values of the reference model update $\Delta\boldsymbol{\hat{g}^{m}_{t+1}}$. The disconnection update gradually reduces the connection parameters between the pill and the rest of the model to $0$, and finally isolates the poison pill from the global model, guaranteeing the attacking effects. 

\parab{Pill Adjustment.}
After the injection,
we use a two-step adjustment to further adjust the pill, improving the generality against multiple detection metrics simultaneously.
In this stage,
we consider two prevailing detection metrics -- distance and cosine similarity. To increase the cosine similarity between the poisoned model update and the benign model update in our method, we balance the magnitudes of both the poison pill's parameters and the other benign parameters.
Similarly, to minimize the distance discrepancy between the poisoned and benign model updates, we adjust the magnitude of the entire poisoned model update, as shown in Figure~\ref{fig:dist adjust}. Thus, we first use the \textbf{similarity-based adjustment}, then use the \textbf{distance-based adjustment}, balancing the \textit{effectiveness} and the \textit{stealthiness} of the poisoned model update. This two-step adjustment is particularly effective when combined with our method, which selectively poisons only a tiny subset of the model’s parameters. By altering just a few parameters, our method preserves a substantial number of benign parameters, which are crucial for making effective adjustments. As a result, the poisoned model update can bypass a wide range of defenses since they are typically designed based on the combination or variants of distance and cosine similarity metrics, and they usually do not anticipate such a focused and minimal interference in the model parameters. 
More details are shown in the extended version.

\section{Evaluation}
\label{sec:evaluation}
This section evaluates how our method enhances existing FL poisoning attacks from three perspectives. First, we assess its \textit{Augmentation Effectiveness} against four poisoning attacks using nine state-of-the-art defenses across three datasets (Section~\ref{subsec:effectiveness}). Second, we examine its \textit{Stealthiness} under two detection metrics (Section~\ref{subsec:metric}). Finally, we conduct a \textit{Generality Analysis}, testing different malicious client proportions, both cross-silo and cross-device settings, and various pill search rules (Section~\ref{subsec:ablation}). Our method substantially strengthens existing attacks, bypassing all nine defenses in over 90\% of cases and raising error rates by up to seven times compared to the originals. It remains effective across diverse data distributions, model architectures, malicious client proportions, and pill search rules.

\subsection{Evaluation Settings}
\label{subsec:experiment setting}
In our experiments, the malicious client proportion is set to $20\%$ by default. We assess 9 standard aggregation rules, including FedAvg, FLTrust, Multi-Krum, Bulyan, Median, Trim, FLDetector, DnC, and Flame, alongside 4 notable model poisoning attacks: sign-flipping, Trim attack, Krum attack, and Min-Max. Experiments utilize 50 clients for MNIST~\cite{lecun1998mnist} and Fashion-MNIST~\cite{fashionmnist}, and 30 for CIFAR-10~\cite{cifar}, accommodating cross-silo and cross-device scenarios. Implementation is done in PyTorch~\cite{paszke2019pytorch}. More details are provided in the extended version.

\begin{figure*}[t]
    \centering
    \includegraphics[width = 0.92\textwidth]{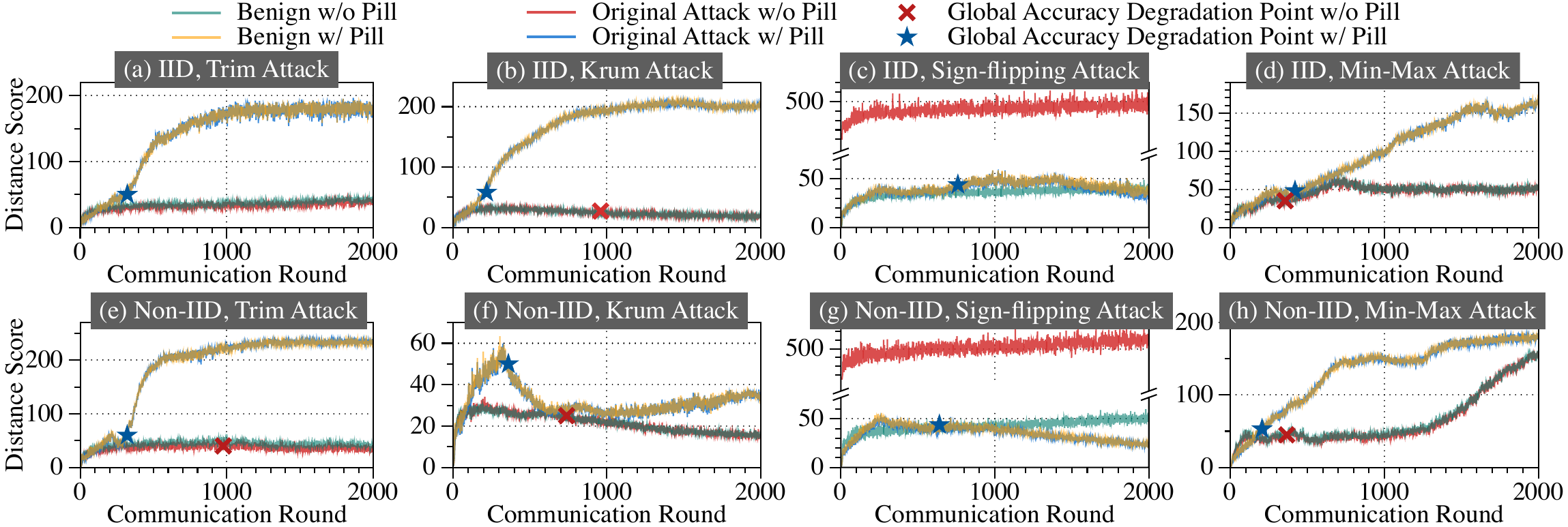}
    \caption{Comparison of Multi-Krum distance score between benign updates and malicious updates when using original poisoning attacks with and without our method.}
    \label{fig:iid distance}
\end{figure*}
\begin{table*}[htbp]
	\centering
        \small
        \setlength{\tabcolsep}{1.8pt}
	    \begin{tabular}{cccccccc|ccccccc}
		\toprule
        \textbf{Data Distribution} & \multicolumn{7}{c}{\textbf{IID}} & \multicolumn{7}{c}{\textbf{Non-IID}}\\
        \otoprule
		\textbf{Attack} & \textbf{FedAvg} & \textbf{FLTrust} & \textbf{MKrum} & \textbf{Bulyan} & \textbf{Median} & \textbf{Trim} & \textbf{FLD} & \textbf{FedAvg} & \textbf{FLTrust} & \textbf{MKrum} & \textbf{Bulyan} & \textbf{Median} & \textbf{Trim} & \textbf{FLD}\\
		\otoprule
		\multirow{1}{*}{No Attack} & 0.106 & 0.104 & 0.103 & 0.108 & 0.127 & 0.107 & 0.116 & 0.111 &  0.119 &  0.113 &  0.113 &  0.140 &  0.114 & 0.123 \\
        \midrule
		\multirow{1}{*}{Sign-Flipping} & \textbf{0.964} & 0.109 & 0.108 & 0.110 & 0.130 & 0.108 & 0.117 & \textbf{0.909} &  0.119 &  0.114 &  0.119 &  0.144 &  0.120 & 0.125 \\
		\cmidrule{2-15}
        \rowcolor{BlueGray2}
        \cellcolor{White} + Poison Pill & 0.320 & \textbf{0.116} & \textbf{0.162} & \textbf{0.151} & \textbf{0.323} & \textbf{0.148} & \textbf{0.699} & 0.269 &  \textbf{0.120} &  \textbf{0.239} &  \textbf{0.164} &  \textbf{0.364} &  \textbf{0.168} & \textbf{0.242}\\
		\midrule
		\multirow{1}{*}{Trim Attack} & 0.112 & 0.111 & 0.111 & 0.115 & 0.132 & 0.114 & 0.116 &  0.125 &  0.115 &  0.121 &  0.125 &  0.153 &  0.122 & 0.122\\
		\cmidrule{2-15}
        \rowcolor{BlueGray2}
        \cellcolor{White} + Poison Pill & \textbf{0.508} & \textbf{0.139} & \textbf{0.334} & \textbf{0.126} & \textbf{0.284} & \textbf{0.127} & \textbf{0.120} & \textbf{0.528} &  \textbf{0.148} &  \textbf{0.455} &  \textbf{0.143} &  \textbf{0.287} &  \textbf{0.146} & \textbf{0.136} \\
		\midrule
		\multirow{1}{*}{Krum Attack} & 0.107 & 0.108 & 0.114 & 0.123 & 0.141 & 0.112 & \textbf{0.668} & 0.116 &  0.117 &  0.124 &  0.138 &  0.173 &  0.122 & 0.410 \\
		\cmidrule{2-15}
        \rowcolor{BlueGray2}
        \cellcolor{White} + Poison Pill & \textbf{0.183} & \textbf{0.118} & \textbf{0.283} & \textbf{0.161} & \textbf{0.362} & \textbf{0.146} & 0.631 & \textbf{0.428} &  \textbf{0.127} &  \textbf{0.280} &  \textbf{0.187} &  \textbf{0.415} &  \textbf{0.182} & \textbf{0.704} \\
        \midrule
        \multirow{1}{*}{Min-Max Attack} & 0.117 & 0.108 & 0.118 & 0.135 & 0.142 & 0.128 & 0.111 & 0.124 &  0.119 &  0.142 &  \textbf{0.166} &  0.162 &  0.145 & 0.136 \\
		\cmidrule{2-15}
        \rowcolor{BlueGray2}
        \cellcolor{White} + Poison Pill & \textbf{0.439} & \textbf{0.129} & \textbf{0.361} & \textbf{0.136} & \textbf{0.343} & \textbf{0.150} & \textbf{0.715} & \textbf{0.521} &  \textbf{0.136} &  \textbf{0.339} &  0.153 &  \textbf{0.368} &  \textbf{0.184} & \textbf{0.335} \\
		\bottomrule
	    \end{tabular}
        \caption{Error rates on Fashion-MNIST under cross-silo setting with 10\% malicious clients in the 50-client FL system.}
	\label{Table:fmnist-cs-10}
\end{table*}
\begin{table*}[t]
	\centering
        \small
        \setlength{\tabcolsep}{2.3pt}
	    \begin{tabular}{cccccccccc|ccccccccc}
		\toprule
        \textbf{Distribution} & \multicolumn{9}{c}{\textbf{IID}} & \multicolumn{9}{c}{\textbf{Non-IID}}\\
        \otoprule
		\textbf{Attack} & \textbf{FAvg} & \textbf{FLT} & \textbf{MKr} & \textbf{Bulyan} & \textbf{Med} & \textbf{Trim} & \textbf{DnC} & \textbf{FLD} & \textbf{Flame} & \textbf{FAvg} & \textbf{FLT} & \textbf{MKr} & \textbf{Bulyan} & \textbf{Med} & \textbf{Trim} & \textbf{DnC} & \textbf{FLD} & \textbf{Flame}\\
		\otoprule
        No Attack & 0.48 & 0.48 & 0.50 & 0.46 & 0.55 & 0.45 & 0.44 & 0.49 & 0.49 & 0.48 & 0.47 & 0.49 & 0.49 & 0.58 & 0.52 & 0.46 & 0.50 & 0.53\\
        \midrule
		Sign-Flipping & \textbf{0.89} & 0.47 & 0.58 & 0.53 & 0.62 & 0.46 & 0.46 & 0.49 & 0.50 & \textbf{0.90} & 0.51 & 0.51 & 0.62 & 0.65 & 0.57 & 0.50 & 0.60 & 0.53\\
		\cmidrule{2-19}
        \rowcolor{BlueGray2}
        \cellcolor{White} + Poison Pill & 0.73 & \textbf{0.88} & \textbf{0.92} & \textbf{0.69} & \textbf{0.70} & \textbf{0.69} & \textbf{0.53} & \textbf{0.89} & \textbf{0.70} & 0.87 & \textbf{0.86} & \textbf{0.89} & \textbf{0.67} & \textbf{0.76} & \textbf{0.68} & \textbf{0.56} & \textbf{0.90} & \textbf{0.67}\\
		\midrule
		Trim ATK & 0.48 & 0.50 & 0.48 & 0.53 & 0.62 & 0.51 & 0.45 & 0.45 & 0.50 & 0.57 & 0.49 & 0.60 & 0.59 & 0.65 & 0.54 & 0.48 & 0.48 & 0.50\\
		\cmidrule{2-19}
        \rowcolor{BlueGray2}
        \cellcolor{White} + Poison Pill & \textbf{0.85} & \textbf{0.87} & \textbf{0.88} & \textbf{0.65} & \textbf{0.67} & \textbf{0.66} & \textbf{0.51} & \textbf{0.89} & \textbf{0.54} & \textbf{0.89} & \textbf{0.86} & \textbf{0.90} & \textbf{0.77} & \textbf{0.68} & \textbf{0.63} & \textbf{0.51} & \textbf{0.89} & \textbf{0.62}\\
		\midrule
		Krum ATK & 0.47 & 0.54 & 0.47 & 0.56 & 0.54 & 0.51 & 0.45 & 0.80 & 0.50 & 0.48 & 0.50 & 0.49 & 0.52 & 0.64 & 0.51 & 0.48 & \textbf{0.89} & 0.50\\
		\cmidrule{2-19}
        \rowcolor{BlueGray2}
        \cellcolor{White} + Poison Pill & \textbf{0.70} & \textbf{0.89} & \textbf{0.90} & \textbf{0.76} & \textbf{0.75} & \textbf{0.64} & \textbf{0.52} & \textbf{0.89} & \textbf{0.87} & \textbf{0.72} & \textbf{0.84} & \textbf{0.90} & \textbf{0.67} & \textbf{0.74} & \textbf{0.64} & \textbf{0.58} & 0.88 & \textbf{0.87}\\
        \midrule
        Min-Max ATK & 0.45 & 0.50 & 0.46 & 0.50 & 0.57 & 0.46 & 0.51 & 0.52 & 0.52 & 0.47 & 0.50 & 0.49 & 0.56 & 0.63 & 0.60 & 0.47 & 0.48 & 0.48\\
		\cmidrule{2-19}
        \rowcolor{BlueGray2}
        \cellcolor{White} + Poison Pill & \textbf{0.75} & \textbf{0.71} & \textbf{0.90} & \textbf{0.77} & \textbf{0.80} & \textbf{0.64} & \textbf{0.54} & \textbf{0.90} & \textbf{0.81} & \textbf{0.66} & \textbf{0.64} & \textbf{0.88} & \textbf{0.67} & \textbf{0.78} & \textbf{0.66} & \textbf{0.52} & \textbf{0.90} & \textbf{0.79}\\
		\bottomrule
	    \end{tabular}
    \caption{Error rates on CIFAR-10 under cross-silo setting with 20\% malicious clients in the 30-client FL system.}
	\label{Table:cifar-cs-20}
\end{table*}

\subsection{Augmentation Effectiveness}
\label{subsec:effectiveness}
This section provides a detailed analysis of our method's augmentation effect on Fashion-MNIST in a 50-client cross-silo FL system with 20\% malicious clients. We test our method on both IID and non-IID data, resulting in an average error rate increase of over 0.25 for all baseline attacks, showing our method's \textit{effectiveness} and \textit{compatibility}.

\parab{Results on IID Data.}
The error rates of four baseline FL poisoning attacks, with and without our method, are shown in the left half of Table~\ref{Table:fmnist-cs-20}. Our method enhances the error rates of the existing poisoning attacks in $23$ out of $28$ scenarios, against FedAvg and five defenses. The maximum increase in error rate is $0.658$, and the average increase reaches $0.274$. This substantial elevation from the attack-free baseline error rate of $0.109$ underscores our method's capability to significantly compromise existing defenses' integrity. 

\parab{Results on Non-IID Data.}
Evaluations on non-IID data further validate the effectiveness of our method, demonstrating its superiority in $23$ of $28$ cases. The highest error rate increase reaches $0.637$, with an average increase of $0.281$. Although there is a slight reduction in the maximal error rate increase in the non-IID setting, these results still demonstrate our method's ability to effectively enhance attacks in more complex and heterogeneous data environments.

All attacks augmented by our method can bypass all baseline defenses, including FLTrust and FLDetector, with the exception of the sign-flipping attack. Notably, the Min-Max attack demonstrates superior effectiveness in non-IID data settings, achieving significant improvements compared to its performance on IID data. Other attacks also exhibit similar error rate improvements relative to their results on IID data, indicating that our method maintains its robustness and effectiveness in more complex data environments. More detailed analyses are presented in the extended version.

\subsection{Stealthiness Analysis}
\label{subsec:metric}
To further assess our method, we examine its \textit{stealthiness} during FL training, focusing on its impact on the distance and cosine similarity scores of existing poisoning attacks. Results show that our method can make malicious clients appear as benign, or even more “benign” than genuine clients, owing to its pill design with distance-based and similarity-based adjustments. As shown in Figure~\ref{fig:iid distance}, the average distance scores of malicious clients (with our method) across four baseline model poisoning attacks closely match or even coincide with those of benign clients throughout training. Detailed results, including similarity score analyses, are provided in the extended version.

\subsection{Generality Analysis}
\label{subsec:ablation}
In this section, we further discuss the \textit{generality} of our method across three key factors: malicious client proportion, client participation frequency, and datasets \& model architectures. The results indicate that our method consistently maintains its effectiveness despite changes in these conditions, demonstrating its adaptability and broad applicability in augmenting existing attacks.

\parab{Impact of The Malicious Client's Proportion.}
We first assess the effectiveness of our method in both IID and non-IID cross-silo FL systems with only $10$\% of clients compromised, as shown in Table~\ref{Table:fmnist-cs-10}. This setup reveals that all baseline model poisoning attacks yield lower error rates on the global model compared with scenarios with $20$\% compromised clients. While the increase in error rates is less than those in the $20$\% compromised client scenario, our method still effectively raises the global model's error rates in $25$/$26$ out of $28$ cases (IID/non-IID setting). The maximum increase in error rates reaches $0.403$, with an average increase of $0.144$. This average is notably higher (>$2$x higher) than the error rates observed in attack-free FL conditions. Specifically, our method helps sign-flipping/Trim/Krum/Min-Max attacks achieve an average error rate increase of $0.133$/$0.094$/$0.136$/$0.209$. More detailed results are presented in the extended version.

\parab{Impact of The Client Participation Frequency.}
We then extend the evaluation of our method to a cross-device FL system, where only $40$\% of clients are selected for participation in each communication round. This setup results in less frequent participation from each client and a fluctuating proportion of malicious clients across different rounds. The maximum error rate increase with our method is $0.639$, with an average increase across different attacks and defenses of $0.279$. These results are consistent with those from the cross-silo FL system, underscoring our method's effectiveness and \textit{generality} across different FL configurations. This evaluation demonstrates our method's robust performance and adaptability, not only in a controlled cross-silo environment but also under the more various conditions in cross-device FL systems. More details are presented in the extended version.

\parab{Impact of The Datasets and Model Architectures.}
Following the evaluation with the Fashion-MNIST dataset, we test our method on the MNIST and CIFAR-10 dataset, employing the four-layer CNN model and the AlexNet model to further verify our method's \textit{generality} across different datasets. 
The collective results show that our method performs even better with larger datasets or more complex machine learning models. This trend confirms the \textit{generality} of our method by revealing its capability to maintain consistent performance enhancements regardless of the dataset or model complexity involved. Specifically, our method helps all four baseline attacks bypass all nine baselines on CIFAR-10 dataset, achieving $0.288$ error rate increase on average, presented in Table~\ref{Table:cifar-cs-20}. More detailed results on MNIST dataset are shown in the extended version.

Beyond the three key factors, we also investigate the impact of the pill search algorithm in the extended version. The results indicate that the ``approximate max pill search'' algorithm outperforms the ``approximate min pill search'' in 41 out of 56 cases ($\sim$73\%), highlighting its effectiveness in selecting the most influential parameters to maximize attack impact. Additional findings on ablation studies and generalizability are also provided in the extended version.

\section{Discussion}
\label{sec:discussion}
To further evaluate the robustness of our method when defenses are aware of the attack strategies (white-box scenario), we design an adaptive defense and present the experimental details in the extended version. Despite the adaptive defense's attempt to incorporate both cosine similarity and distance metrics, it remains insufficient to thwart the enhanced capabilities of our method. We also presented a detailed discussion of the limitations and future directions in the extended version.

\section{Conclusion}
\label{sec:conclusion}
We propose a novel attack-agnostic augmentation method to enhance FL poisoning attacks by concentrating them into a tiny subnet (\ie, \textit{pill}). Our approach comprises \textit{pill construction}, \textit{pill poisoning}, and \textit{pill injection}, enabling existing FL poisoning attacks to achieve over $2\times$ higher error rates on average. By intensifying the inherent vulnerabilities in current FL defenses, our method underscores the urgent need for more robust and fine-grained detection mechanisms.

\section*{Acknowledgements}
We thank the anonymous reviewers for their valuable feedback.
This work was supported in part by the National Natural Science Foundation of China (NO. 62472284, 62572426), Shanghai Key Laboratory of Scalable Computing and Systems.
The work of H. Wang was supported in part by the United States National Science Foundation (NSF) under grants 2534286, 2523997, 2315612, and 2332638.  

\bibliography{reference}

\input{sections/appendix}

\end{document}

%% file: sections/appendix.tex
\appendix
\label{appendix}

\section{Additional Details of Existing Attacks and Defenses}
\label{appendix:baseline}
\subsection{Additional Details of Existing Attacks in FL}
\label{appendix:additional attack info}
Although model poisoning attacks are effective, existing attacks have limited \textit{stealthiness} and can be detected by many existing defenses. Our goal is hence to demonstrate that such attacks can be augmented in a uniform way. 
Model poisoning attacks  directly manipulate the parameters uploaded by clients, with a minimal interference to the local training process. Among these attacks, the simplest form is the sign-flipping attack, which directly flips the model update and scales it with a constant factor. 
A-Little-is-Enough~\citep{baruch2019little} generates malicious updates within a calculated perturbation range to deceive the global model.
Adaptive attacks~\citep{fang2020local}, such as the Trim attack and the Krum attack, dynamically scale malicious updates based on parameter values and distances. The Min-Max and Min-Sum Attacks~\citep{shejwalkar2021manipulating} provide a dynamic scaling for malicious updates based on different distance-based criteria. MPAF~\citep{cao2022mpaf} aims to drive the global model towards a predefined target model with poor performance on given FL tasks. For the sake of generality, we employ the sign-flipping attack, two types of adaptive attacks~\citep{fang2020local} (the Trim attack and the Krum attack), and the Min-Max attack~\citep{shejwalkar2021manipulating} as the baseline attacks in this paper.

\subsection{Additional Details of Existing Defenses in FL}
\label{appendix:additional defense info}
\textit{Adaptive Client Filtering.} 
These techniques such as Krum and Multi-Krum~\citep{krum} filter out malicious clients through single or multiple rounds of client selection based on distance scores. FLTrust~\citep{cao2020fltrust} computes trust scores using the cosine similarity between each client update and the server model’s update for weighted averaging. SignGuard~\citep{xu2021signguard} employs sign-based clustering combined with norm-based thresholding to identify and filter malicious clients. Flame~\citep{nguyen2022flame} and Deepsight~\citep{rieger2022deepsight} propose adaptive clustering and clipping to safeguard against backdoor attacks. SkyMask~\citep{yan2023skymask} clusters trainable feature masks of clients to assess each client’s risk level.

\textit{Statistical Parameter Aggregation.} 
Approaches like Median and Trim~\citep{yin2018trim} use coordinate-wise median or trimmed mean values to aggregate model updates. Bulyan~\citep{bulyan} enhances robustness by integrating Krum with Trim techniques. Fool's Gold~\citep{foolsgold} applies an adaptive learning rate based on inter-client contribution similarity to mitigate the effects of malicious updates. SparseFed~\citep{panda2022sparsefed} aggregates sparsified updates, reducing the risk of model poisoning attacks.

\textit{Client-dominant Detection.} 
Siren~\citep{guo2021siren} and Siren$^+$~\citep{guo2024siren+} set proactive accuracy-based alarms at the client level with the corresponding server-side decisions to counter various model poisoning attacks. FL-WBC~\citep{sun2021flwbc} introduces client-side noise to diminish the efficacy of attacks and shorten their duration. FLIP~\citep{zhang2023flip} achieves higher robustness through client-side reverse-engineering defenses against extensive poisoning strategies. LeadFL~\citep{zhu2023leadfl} uses a client-side Hessian matrix optimization to reduce the impact of adversarial patterns on backdoor and targeted attacks.

\textit{Other Advanced Metrics and Pipelines.} 
Various studies employ other sophisticated metric pipelines designed for detection to ensure robust defense against poisoning attacks. These include techniques proposed in studies such as Zeno~\citep{xie2019zeno}, CRFL~\citep{xie2021crfl}, FedRecover~\citep{cao2023fedrecover}, FLCert~\citep{cao2022flcert}, FLDetector~\citep{zhang2022fldetector}, and MESAS~\citep{krauss2023mesas}.

Here are more details of the baseline defenses used in our paper:

\parab{Krum and Multi-Krum (MKrum)~\citep{krum}.}
Krum uses a distance score as the metric. In each round, the Krum server sums the distances between each client update $\boldsymbol{g}_t^{(i)}$ and its $K-m-2$ neighbors, and uses these sums as the scores for all the clients. The Krum server then selects the client’s model update with the lowest score. Multi-Krum is a variant of Krum that uses iterative Krum to pick multiple candidates for aggregation.

\parab{Coordinate-wise Median (Median)~\citep{yin2018trim}.}
Coordinate-wise Median (Median) uses the per-parameter median values of the model updates from the clients as the aggregated global model update, which is then used to generate the next-round global model.

\parab{Trimmed Mean (Trim)~\citep{yin2018trim}.}
Trimmed Mean (Trim) calculates per-parameter trimmed mean values of the client model updates and packs them as the global model update.

\parab{Bulyan~\citep{bulyan}.}
Bulyan is a combination of Krum and Trim. It first uses the Krum-based method to select multiple candidates, and uses the per-parameter trimmed mean values of the candidate model updates as the final global model update.

\parab{FLTrust~\citep{cao2020fltrust}.}
FLTrust trains a server model with a small root dataset. In each round, it computes the clipped cosine similarities between the server model update and client updates as trust scores, and then uses the trust scores as weights to aggregate all the normalized client model updates.

\parab{FLDetector (FLD)~\citep{zhang2022fldetector}.}
FLDetector filters out malicious clients by checking the multi-round consistency of all client updates. Malicious updates typically have lower consistency compared to benign ones.

\parab{Flame~\citep{nguyen2022flame}.}
Flame utilizes HDBSCAN-based~\citep{hdbscan} dynamic clustering to filter out malicious clients, and aggregates median-clipped benign updates with adaptive noise as the global model update.

\begin{table}[t]
\setlength{\tabcolsep}{10pt}
\footnotesize
\centering
    \begin{threeparttable}
                \begin{tabular}{ccc}
                    \toprule
                    \textbf{Layer Type} & \Centerstack{\textbf{Original} \\\textbf{CNN Model}}                & \Centerstack{\textbf{Our} \\\textbf{Pill Blueprint}}                                   \\\otoprule
                    Input & $28\times28\times1$ & $28\times28\times1$\\
                    Conv2d & $3\times3\times30$ &  $3\times3\times1$\\
                    ReLU & -\tnote{1} & - \\
                    MaxPool2d & $2\times2$ & $2\times2$ \\
                    Conv2d & $3\times3\times50$ &  $3\times3\times1$\\
                    ReLU & - & - \\
                    MaxPool2d & $2\times2$ & $2\times2$ \\
                    Linear & $1250\times100$ & $25\times10$\\
                    ReLU & - & - \\
                    Linaer & $100\times 10$ & $10$\\ 
                    Softmax & - & $\times\tnote{2}$\\
                    \bottomrule
                \end{tabular}
            \begin{tablenotes}
                \item[1] ``-'' represents that the model has this layer with no specified configuration.
                \item[2] ``$\times$'' represents that the model does not contain this layer.
            \end{tablenotes}\
    \end{threeparttable}
\caption{Architectures of the original CNN model and the corresponding pill blueprint.}
\label{tab:model}
\end{table}
\begin{table}[t]
\centering
\footnotesize
\setlength{\tabcolsep}{10pt}
    \begin{threeparttable}
                \begin{tabular}{ccc}
                    \toprule
                    \textbf{Layer Type} & \Centerstack{\textbf{Original} \\\textbf{AlexNet}}                & \Centerstack{\textbf{Our} \\\textbf{Pill Blueprint}}                                   \\\otoprule
                    Input & $32\times32\times3$ & $32\times32\times3$\\
                    Conv2d & $11\times11\times48$ & $11\times11\times1$ \\
                    ReLU & - & - \\
                    MaxPool2d & $3\times 3$ & $3\times 3$ \\
                    Conv2d & $3\times3\times96$ & $3\times3\times1$ \\
                    ReLU & - & - \\
                    MaxPool2d & $3\times 3$ & $3\times 3$ \\
                    Conv2d & $3\times3\times192$ & $3\times3\times1$ \\
                    ReLU & - & - \\
                    Conv2d & $3\times3\times192$ & $3\times3\times1$ \\
                    ReLU & - & - \\
                    Conv2d & $3\times3\times128$ & $3\times3\times1$ \\
                    ReLU & - & - \\
                    MaxPool2d & $3\times 3$ & $3\times 3$ \\
                    Linear & $4608\times 1024$ & $36\times1$ \\
                    ReLU & - & - \\
                    Linear & $1024\times 512$ & $1\times10$ \\
                    ReLU & - & - \\
                    Linear & $512\times 10$ & $10$ \\
                    Softmax & - & $\times$\\
                    \bottomrule
                \end{tabular}
            \begin{tablenotes}
                \item[1] ``-'' represents that the model has this layer with no specified configuration.
                \item[2] ``$\times$'' represents that the model does not contain this layer.
            \end{tablenotes}
    \end{threeparttable}
\caption{Example architectures of original AlexNet and the corresponding pill blueprint.}
\label{tab:alexnet}
\end{table}

\begin{table}[t]
  \centering
    \footnotesize
    \centering
    \rowcolors{13}{BlueGray2}{BlueGray2}
    \begin{tabular}{cc}
                    \toprule
                    \textbf{Symbol}                & \textbf{Meaning}                                   \\\otoprule%
                   $T$ & Total FL communication round \\
                   $t$ & FL communication round index \\
                   $K$ & Total client number \\
                   $\boldsymbol{g}$ & Global model of the FL training \\
                   $\boldsymbol{g}_t$ & Global model in round $t$ \\
                   $lr$ & Learning rate \\
                   $f()$ & Loss function used in the FL training \\
                   \midrule
                   $i$ & Client index \\
                   $E$ & Local training epoch number \\
                   $D^{(i)}$ & Local training data on client $i$ \\
                   $\Delta\boldsymbol{g}_t^{(i)}$ & Local model update of client $i$ in round $t$ \\
                   \midrule
                   $m$ & Total amount of malicious clients \\
                   $D^m$ & Aggregated data from compromised clients\\
                   $\Delta\hat{\boldsymbol{g}}_t^{m}$ & Update of extra-trained model in round $t$\\
                   $\Delta\boldsymbol{\widetilde{g}}_{t}$ & Estimated global model update in round $t$ \\
                   $\Delta\boldsymbol{g}_t^{zero}$ & Disconnection update in round $t$\\
                   $M$ & Selected malicious subnetwork \\
                   $M_{disc}$ & Disconnection mask corresponding to $M$ \\
                   $C_{iter}$ & Max malicious update adjustment iteration\\
                   $C_{\uparrow}$ & Up-scaling factor \\
                   $C_{\downarrow}$ & Down-scaling factor\\
                    \bottomrule
                \end{tabular}
                \captionof{table}{Main notations. Symbols in the gray part are used for attacks.}
                \label{table:symbols}
\end{table}%

\section{Detailed Threat Model}
\label{appendix:threat model}
\parab{\ul{Attacker's Goal and Capabilities.} }
This paper focuses on improving the effectiveness of existing poisoning attacks in FL. Similar to previous work~\citep{fang2020local,shejwalkar2021manipulating}, an attacker aims to raise the error rates of the global model on a specific class or multiple classes by sending poisoned model updates via compromised clients during the iterative aggregation.
Our method does not require any additional knowledge compared with existing FL poisoning attacks. Hence we reuse the typical threat model in existing studies~\citep{fang2020local,shejwalkar2021manipulating}.
The attacker has a complete control of the compromised clients, including their local data, local training, and uploading process. With the aggregated resources on the compromised clients, the attacker may aggregate the local data from the compromised clients to do extra training or aggregate their local updates to do model estimation. 
The attacker may or may not know the updates of other benign clients, depending on the confidentiality of the communication channels between the server and clients.
Besides, the attacker cannot access the server's information, including the aggregation rules or the selected clients in each round.

\parab{\ul{Defense Settings.} }
Most of the defenses in FL are deployed and executed on the server. We adopt a similar defense setting as existing studies~\citep{krum,yin2018trim,cao2020fltrust,guo2021siren}. 
The server cannot directly analyze the local data or the local training of clients. It can only detect malicious clients through model updates from different clients. The server can collect and possess a root test dataset to provide more accurate and robust detection, while the data of such a root test dataset cannot be derived from clients. The data distribution of this root test dataset may or may not be the same as the data distribution across the clients.

\begin{algorithm}[t]
    \caption{\textbf{Our method's workflow. Assume the first $m$ clients as malicious clients. }}
    \label{alg:sys}
    \footnotesize
        \Function{\textcolor{AlgRed}{MalUpdate}($i$,$t$,$\boldsymbol{g}_t$)}{
            \Comment{1.Pill Construction}
            $\boldsymbol{M}$, $\boldsymbol{M}_{disc} \assign$ \FuncCall{Search}{$\boldsymbol{g}_t$}\;
            \Comment{2.Pill Poisoning}
            $\boldsymbol{\hat{g}}^{m}_{t+1} \assign \boldsymbol{g}_t$\;
            \For{each epoch $e_{extra} \assign 1,\cdots,E_{extra}$}{
                \texttt{sample} $B^{m}$ \texttt{from aggregated local data $D^{m}$ on compromised clients} \;
                $\boldsymbol{\hat{g}}^{m}_{t+1} \assign \boldsymbol{\hat{g}}^{m}_{t+1} - lr\cdot \nabla f(B^m,\boldsymbol{\hat{g}}^{m}_{t+1}$)\;
            }
            $\Delta\boldsymbol{\hat{g}^{m}_{t+1}}\assign \boldsymbol{g}_t - \boldsymbol{\hat{g}}^{m}_{t+1}$\;
            $\Delta\boldsymbol{g}_{t+1}^{(i)} \assign$ \FuncCall{Poisoning}{$i$,$t$,$\boldsymbol{param}$,$\Delta\boldsymbol{\hat{g}}^{m}_{t+1}$}\;
            $\Delta\boldsymbol{g}_{t+1}^{(i)} \assign \boldsymbol{M}\odot\Delta\boldsymbol{g}_{t+1}^{(i)}$\;
            \Comment{3.Pill Injection}
            $\mathop{param}\assign\{\Delta\boldsymbol{g'}^{(1),\cdots,(m)}_{t+1}\}$\;
            $\Delta\boldsymbol{\widetilde{g}}_{t+1} \assign$ \FuncCall{Estimation}{$i$,$t$,$\boldsymbol{g}_t$,$\mathop{param}$}\;
            $\Delta\boldsymbol{g}_{t+1}^{(i)} \assign \Delta\boldsymbol{g}_{t+1}^{(i)} + (\mathbf{1}-\boldsymbol{M})\odot\boldsymbol{\Delta\widetilde{g}}_{t+1}$\;
            $\Delta\boldsymbol{g}_{t+1}^{zero}\assign \mathbf{0}-\boldsymbol{g}_t$\;
            $\Delta\boldsymbol{g}_{t+1}^{(i)} \assign \boldsymbol{M}_{disc}\odot\Delta\boldsymbol{g}_{t+1}^{zero} + (\mathbf{1}-\boldsymbol{M}_{disc})\odot\Delta\boldsymbol{g}_{t+1}^{(i)}$\;
            $\mathop{param} = \mathop{param} \bigcup \{\boldsymbol{M}_{all} = M+M_{disc}\}$\;
            $\Delta\boldsymbol{g}_{t+1}^{(i)}\assign$\FuncCall{SimAdjust}{$\mathop{param}$,$\boldsymbol{\Delta\widetilde{g}}_{t+1}$,$\Delta\boldsymbol{g}_{t+1}^{(i)}$}\;
            $\Delta\boldsymbol{g}_{t+1}^{(i)}\assign$\FuncCall{DistAdjust}{$\mathop{param}$,$\boldsymbol{\Delta\widetilde{g}}_{t+1}$,$\Delta\boldsymbol{g}_{t+1}^{(i)}$}\;
            \Return{$\Delta\boldsymbol{g}^{(i)}_{t+1}$}
        }
\end{algorithm}

\section{Main Notations and Pseudo-Code}
\label{appendix:notation}
Table~\ref{table:symbols} presents the main notation used in the paper,W and Algorithm~\ref{alg:sys} shows the pseudo-code of our proposed attack augmentation method.

\section{Additional Details of Concrete Pill Blueprints}
\label{appendix:alexnet}

Table~\ref{tab:model} and Table~\ref{tab:alexnet} illustrate the model structures of the CNN model and the simplified AlexNet, with their corresponding pill's blueprints.

\section{Detailed Pill Search Algorithm}
\label{appendix:pill search}

\begin{figure*}[ht]
    \centering
    \includegraphics[width=0.98\textwidth]{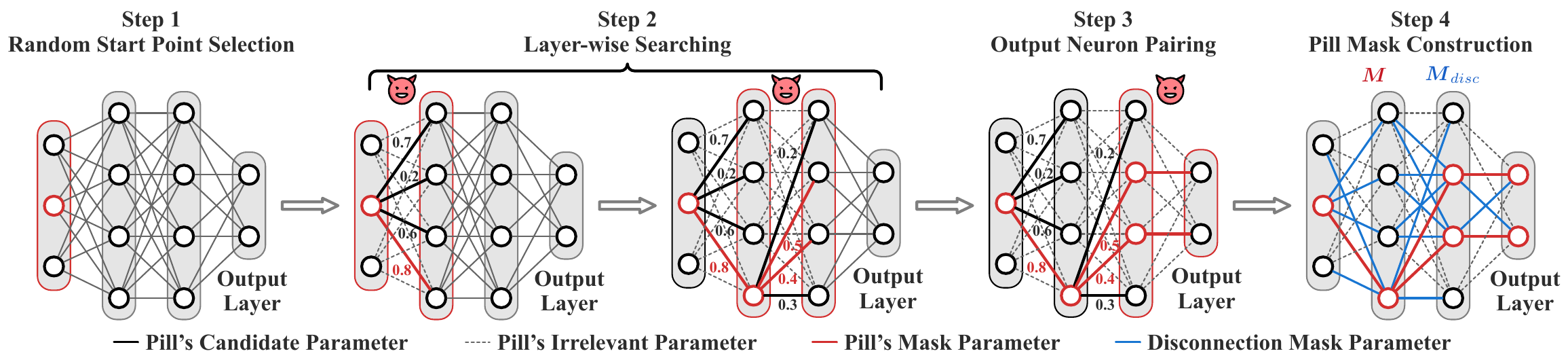}
    \caption{An example of the ``approximate max pill search'' algorithm in our augmentation method.}
    \label{fig:max search}
\end{figure*}

A complete procedure of the ``approximate max pill search'' algorithm consists of the following four steps. To improve readability, we use ``neuron'' to represent both neurons in fully connected layers and channels in convolutional layers, and we use the classification task as an example:
\begin{description}[leftmargin=0cm]
    \item[\ul{Step 1 Random Start Point Selection:}] At the beginning of the search, we randomly choose a subset of neurons from the first layer $l_1$ of the target model, based on the structure of the first layer $l^p_1$ in the pill's blueprint and its neuron number $\mathcal{N}^p_1$. The selected neurons are termed as $\mathcal{V}_1$, and defined as start points, which are then fixed across the entire FL training.
    \item[\ul{Step 2 Layer-wise Search:}] For each subsequent layer $l_i$ in the target model, we first calculate the sum of the weights from the selected neurons $\mathcal{V}_{i-1}$ in layer $l_{i-1}$ to each neuron in $l_i$. Then, we rank all the neurons in $l_i$ based on the parameter value sums and choose top $\mathcal{N}^p_i$ neurons in $l_i$ as the new $\mathcal{V}_i$, where $\mathcal{N}^p_i$ represents the number of neurons in the $i$th layer of pill's blueprint. $\mathcal{V}_i$ and all the parameters from $\mathcal{V}_{i-1}$ to $\mathcal{V}_i$ are recorded.
    \item[\ul{Step 3 Output Neuron Pairing:}] After visiting $l_{L-1}$ layers, where $L$ is the total number of layers in the target model, $||\mathcal{V}_{L-1}||$ should equal to the neuron number in $l_{L}$ of the target model, which also equals to the number of classes. We select all the neurons in the target model's $l_{L}$ layer into our pill to construct $\mathcal{V}_{L}$. Then, we only record the parameters from one neuron in $\mathcal{V}_{L-1}$ to only one neuron in $\mathcal{V}_{L}$ based on the index order (\ie, the first neuron in $\mathcal{V}_{L-1}$ is paired with the first neuron in $\mathcal{V}_{L}$). 
    Since $||\mathcal{V}_{L-1}||$ equals to $||\mathcal{V}_{L}||$, the number of recorded parameters equals to the number of classes, 
    avoiding poisoning too many parameters in a single layer.
    \item[\ul{Step 4 Pill Mask Construction:}]
    With the recorded $\mathcal{V}_{i}$ and the corresponding parameters, we construct two masks $\boldsymbol{M}$ and $\boldsymbol{M}_{disc}$. The mask $\boldsymbol{M}$ records the pill's parameters in the target model, and the disconnection mask $\boldsymbol{M}_{disc}$ records the parameters of the connections between the pill and the rest of the target model. $\boldsymbol{M}$ is used for poisoning,
    while $\boldsymbol{M}_{disc}$ is used to disconnect the poison pill from the target model, maintaining the integrity and performance of the pill during poisoning.
    The two masks have the same shape as the target model's parameters.
    To construct $\boldsymbol{M}$, we set the locations corresponding to the pill parameters to $1$, and the others to $0$. 
    Based on $\boldsymbol{M}$, we can similarly obtain the corresponding disconnected mask $\boldsymbol{M}_{disc}$, which sets the locations corresponding to the parameters from neurons other than $\mathcal{V}_{i-1}$ to $\mathcal{V}_i$ in each model layer $l_i$ (except for the $L$th layer since we choose all the neurons from it), and also those corresponding to parameters from $\mathcal{V}_{i-1}$ to other pill irrelevant neurons in each model layer $l_i$ 
    to $1$.  
    The two masks are used in the Pill Injection Stage.
\end{description}
\smallskip
\parab{\ul{Example}.} 
Figure~\ref{fig:max search} presents a concrete example of the search algorithm in a 4-layer linear model. Since the start point is randomly selected by the attacker, defense methods can hardly guess it without any prior knowledge. 
In the example, suppose the model is a 4-layer linear model for a binary classification task. Then the pill blue print contains one neuron in the first two layers, and contains two neurons in the last two layers. Initially, we randomly select a start neuron, specifically the second neuron in the first layer in the example. Then, we conduct layer-wise searching when visiting the second and third layers, selecting the parameters with the highest magnitudes. At the forth layer (output layer), we pair the two output neurons with the two selected neurons in the third layer based on the index order. And we finally construct two pill-related masks accordingly.

With the search algorithm, we also reduce the complexity of the pill search from $\mathcal{O}(\prod_1^L \mathcal{N}_i)$ to $\mathcal{O}(\sum_1^L \mathcal{N}_i\cdot\mathcal{N}^p_i)$, where $\mathcal{N}_i$ represents the neuron number of the target model in layer $i$, and $\mathcal{N}^p_i\ll\mathcal{N}_i$ for all the hidden layers. The computational complexity of our pill search is hence much smaller than the computational complexity of one round local training.

\begin{table}[t]
    \small
    \setlength{\tabcolsep}{10pt}
    \begin{center}
        {
                \begin{tabular}{cl}
                    \toprule
                    \textbf{Notation} & \textbf{Description} \\\otoprule
                    $\pattern_1$ & All layers use the adaptive searching strategy \\
                    \midrule
                    $\pattern_2$ & All layers use the one-time searching strategy \\
                    \midrule
                    $\pattern_3$ & \Centerstack[l]{\textit{FE} layers use the adaptive searching strategy \\ \textit{CLS} layers use the repeated searching strategy} \\
                    \midrule
                    $\pattern_4$ & \Centerstack[l]{\textit{FE} layers use repeated searching strategy \\ \textit{CLS} layers use the adaptive searching strategy} \\
                    \midrule
                    $\pattern_5$ & \Centerstack[l]{\textit{FE} layers use the adaptive searching strategy \\ \textit{CLS} layers use the one-time searching strategy} \\
                    \midrule
                    $\pattern_6$ & \Centerstack[l]{\textit{FE} layers use the one-time searching strategy \\ \textit{CLS} layers use the adaptive searching strategy} \\
                    \bottomrule
                \end{tabular}
        }
        \caption{Different dynamic patterns in our method, utilizing along with ``approximate max pill search''.}
        \label{tab:dynamic}
    \end{center}
\end{table}

To make the pill search process dynamic, we also design several patterns to adaptively determine whether to change the pill in the training period, shown in Table~\ref{tab:dynamic} (\textit{FE} represents the convolutional layers, \textit{CLS} represents the linear layers). For more details about each specific dynamic pattern, please refer to Appendix~\ref{appendix:dynamic}. The combination of the ``approximate max pill search'' with different dynamic patterns constructs the complete dynamic pill search, considering both the \textit{stealthiness} and the \textit{efficiency}.

\section{Additional Details of The Dynamic Patterns in Our Method}
\label{appendix:dynamic}
We first design three searching strategies, including one-time searching strategy, repeated searching strategy, and adaptive searching strategy.

In the one-time searching strategy, we search the pill based on the initial global model using the ``approximate max pill search'' algorithm introduced in \S~\ref{subsec:select}, and keep this pill unchanged in the whole FL training. The one-time searching strategy benefits the formation of pill in the global model, while the initial pill may be less effective with the increasing training rounds of the global model, due to the changing importance of the model parameters. 

On the contrary, the repeated searching strategy runs the `approximate max pill search'' algorithm in every training round. The repeated searching strategy can help our method modify more parameters in the global model, and make the pill less traceable. While the attacking effects may be reduced due to the constantly changing pill.

Considering advantages and disadvantages of both the one-time searching strategy and the repeated searching strategy, we design a more flexible searching strategy, termed as ``adaptive searching strategy''. In the adaptive searching strategy, our method searches the new pill only when the pill is not successfully injected into the global model in the last round. The condition:
\begin{equation}
    \texttt{\bfseries Sim(} \boldsymbol{M}\odot\Delta\boldsymbol{g}_t,  \boldsymbol{M}\odot\Delta\boldsymbol{g}_{t}^{(i)}\texttt{\bfseries )} < C_{search}
\end{equation}
should be satisfied to trigger the new subnetwork searching on malicious client $i$, where $C_{search}$ is set as $0.94$ in the experiments. The adaptive searching strategy is a more moderate version of repeated searching.

Since the three searching strategies have their unique advantages, we investigate different combinations of them in the experiments. We further divide the neural network into \textit{Feature Extractor (FE)} and \textit{Classifier (CLS)}. Refer to the CNN model we used, the convolutional layers are regarded as \textit{FE}, and the linear layers are regarded as \textit{CLS}. We use different searching strategies in \textit{FE} and \textit{CLS}, respectively. In all the nine combinations, we test and keep six of them, noted as \textsc{Pattern}$_1$ to \textsc{Pattern}$_6$, shown in Table~\ref{tab:dynamic} in \S~\ref{subsec:select}. Such six patterns construct the entire dynamic pattern set used in our method.

\section{Additional Details of Pill Adjustment}
\label{appendix:pill adjust}
\begin{algorithm*}[htbp]
    \caption{\textbf{Similarity-based and distance-based adjustment functions in the \textbf{Poison Pill Injection} stage.}}
    \label{alg:malfunc}
    \footnotesize
    \begin{multicols}{2}
            \Function{SimAdjust($\mathop{param}$,$\boldsymbol{\Delta\widetilde{g}}_{t+1}$,$\Delta\boldsymbol{g}_{t+1}^{(i)}$)}{
            $\{\Delta\boldsymbol{g'}^{(1),\cdots,(m)}_{t+1}, \boldsymbol{M}_{all}\}\assign\mathop{param}$\;
            $S_{max}\assign \mathop{max}(0, \mathop{max}\{$\FuncCall{Sim}{$\Delta\boldsymbol{\widetilde{g}}_{t+1}$,$\Delta\boldsymbol{g}_{t+1}^{(i)\prime}$}$;i\in\{1,\cdots, m\}\})$\;
            $\mathop{iter}\assign 0$\;
            \While{\FuncCall{Sim}{$\Delta\boldsymbol{\widetilde{g}}_{t+1},\Delta\boldsymbol{g}_{t+1}^{(i)}$}<$S_{max}$ \&\& $\mathop{iter}<C_{iter}$}{
                \If{$\mathop{iter}\% 2$}{
                    $\Delta\boldsymbol{g}_{t+1}^{(i)}\assign (C_{\uparrow}\cdot(\mathbf{1}-\boldsymbol{M}_{all})+\boldsymbol{M}_{all})\odot\Delta\boldsymbol{g}_{t+1}^{(i)}$\;
                    $\mathop{iter}\assign\mathop{iter}+1$\;
                }
                \Else{
                    $\Delta\boldsymbol{g}_{t+1}^{(i)}\assign ((\mathbf{1}-\boldsymbol{M}_{all})+C_{\downarrow} \cdot \boldsymbol{M}_{all})\odot\Delta\boldsymbol{g}_{t+1}^{(i)}$\;
                    $\mathop{iter}\assign\mathop{iter}+1$\;
                }
            }
            \Return{$\Delta\boldsymbol{g}_{t+1}^{(i)}$}\;
        }
        \Function{DistAdjust($\mathop{param}$,$\boldsymbol{\Delta\widetilde{g}}_{t+1}$,$\Delta\boldsymbol{g}_{t+1}^{(i)}$)}{
            $\{\Delta\boldsymbol{g'}^{(1),\cdots,(m)}_{t+1}, \boldsymbol{M}_{all}\}\assign\mathop{param}$\;
            $\mathop{Dist}_{max} \assign \mathop{max}\{||\Delta\boldsymbol{g'}^{(i)}_{t+1} - \boldsymbol{\Delta\widetilde{g}}_{t+1}||;i\in\{1,\cdots,m\}\}$\;
            $\mathop{Dist} \assign ||\Delta\boldsymbol{g}_{t+1}^{(i)} - \boldsymbol{\Delta\widetilde{g}}_{t+1}||$\;
            \If{$||C_{\downarrow}\cdot\Delta\boldsymbol{g}_{t+1}^{(i)} - \boldsymbol{\Delta\widetilde{g}}_{t+1}|| < ||C_{\uparrow}\cdot\Delta\boldsymbol{g}_{t+1}^{(i)} - \boldsymbol{\Delta\widetilde{g}}_{t+1}||$}{
                $C_{dist} \assign C_{\downarrow}$\;
            }
            \Else{
                $C_{dist} \assign C_{\uparrow}$\;
            }
            \While{$\mathop{Dist}\geq\mathop{Dist}_{max}$ \&\& $||C_{dist}\cdot\Delta\boldsymbol{g}_{t+1}^{(i)} - \boldsymbol{\Delta\widetilde{g}}_{t+1}||\leq\mathop{Dist}$}{
                $\Delta\boldsymbol{g}_{t+1}^{(i)} \assign C_{dist}\cdot\Delta\boldsymbol{g}_{t+1}^{(i)}$\;
                $\mathop{Dist} \assign ||\Delta\boldsymbol{g}_{t+1}^{(i)} - ||$\;
            }
            \Return{$\Delta\boldsymbol{g}_{t+1}^{(i)}$}\;
        }
    \end{multicols}
    \vspace{5pt}
\end{algorithm*}
The details of the two pill adjustment methods are presented as follows:

\parab{\ul{Similarity-based Adjustment.}}
As shown in Line \texttt{1--12} of Algorithm~\ref{alg:malfunc}, we first compute the maximum cosine similarity $S_{max}$ between the normal model updates from the compromised clients and the estimated global model update in the current round. Then, we iteratively and alternately reduce the magnitude of the poison pill's parameters with the down-scaling factor $C_{\downarrow}$, and increase the magnitude of the rest estimated global model update's parameters with the up-scaling factor $C_{\uparrow}$, until the cosine similarity between the entire poisoned model update and the estimated global model update is greater than $S_{max}$ or the adjustment total iteration is greater than the threshold $C_{iter}$.

\parab{\ul{Distance-based Adjustment.}}
In the \textbf{Distance-based Adjustment} (Line \texttt{13--24} of Algorithm~\ref{alg:malfunc}), we reuse the up-scaling factor $C_{\uparrow}$ and the down-scaling factor $C_{\downarrow}$ to adjust the magnitude of the entire poisoned model update. The intuition behind the \textbf{Distance-based Adjustment} is shown in Figure~\ref{fig:dist adjust}. We first calculate the maximum distance between the normal model updates from the compromised clients and the estimated global model update in the current round. We use this maximum distance $Dist_{max}$ as the threshold in the distance-based adjustment. Then, we further determine the scaling factor that should be used by applying the two scaling factors $C_{\uparrow}$ and $C_{\downarrow}$ separately to the poisoned model update $\Delta\boldsymbol{g}_{t+1}^{(i)}$. The scaling factor that reduces the distance between $\Delta\boldsymbol{g}_{t+1}^{(i)}$ and $\boldsymbol{\Delta\widetilde{g}}_{t+1}$ is chosen as the initial scaling factor in the subsequent iterative scaling. We stop the scaling until the distance between the $\Delta\boldsymbol{g}_{t+1}^{(i)}$ and $\boldsymbol{\Delta\widetilde{g}}_{t+1}$ is smaller than $Dist_{max}$, or the scaling factor begins to increase such distance (reach the limit of the scaling).

\section{Additional Experimental Configurations}
\label{appendix:exp setup}
\parab{\ul{Model, Dataset, and Hyper-Parameters.}}
In our experiments, we employ a four-layer Convolutional Neural Network (CNN) and a simplified version of AlexNet~\citep{alexnet}. The structures of the models and their corresponding pill blueprints are detailed in Appendix~\ref{appendix:alexnet}. We evaluate our method on three widely-used datasets: MNIST~\citep{lecun1998mnist}, Fashion-MNIST~\citep{fashionmnist}, and CIFAR-10~\citep{cifar}. We use the CNN model on MNIST and Fashion-MNIST datasets, and the AlexNet~\citep{alexnet} on CIFAR-10 dataset. Each experiment is repeated five times to ensure reliability, with the mean and standard deviation (std) of the results reported.

\parab{\ul{IID and Non-IID Data Settings.}}
Our method was assessed under both IID and non-IID data distributions to understand its performance across data heterogeneity. For IID data setting, we uniformly split all the training data into $K$ shards, and distribute each shard to a random client. For non-IID data setting, we utilize the non-IID degree $p$ as defined in prior studies~\citep{fang2020local,guo2021siren}. A higher $p$ indicates greater data heterogeneity among the clients. Specifically, when $p=0.1$, the data configuration is essentially IID. We set $p=0.5$ to to intensify the non-IID condition, under which we create and allocate $K$ non-IID data shards to all the clients, simulating a more realistic and challenging FL environment. Given that FLTrust necessitates a root dataset at the server, we select this dataset first from the available training data. Subsequently, we distribute the remaining data among the clients according to the aforementioned IID and non-IID rules. This approach ensures that there is no overlap between server's data and client's data.

\parab{\ul{Configurations of Dynamic Patterns.}}
As outlined in Section~\ref{subsec:select}, we design six dynamic patterns for the pill search. We systematically evaluate all six patterns and present the results of the most effective strategy.

\parab{\ul{Evaluation Metrics.}}
We use \textit{error rates} -- defined as the proportion of incorrect predictions -- to evaluate attack effectiveness. Given that the model poisoning attacks discussed are all untargeted, higher error rates indicate more effective attacks. To assess the \textit{stealthiness} of our method in delivering malicious updates, we employ two metrics: 1) \textit{cosine similarity score}, measuring alignment with the server's model update in FLTrust; 2) \textit{distance score}, used in Multi-Krum to evaluate the closeness of poisoned updates to benign updates. 

\section{Additional Augmentation Performance Analysis}
\label{appendix:augmentation analysis}
Following are individual improvements of our method on different baseline attacks in IID data setting:
\begin{itemize}
    \item Sign-flipping attack: Its original version achieves a high error rate due to its aggressive and brute design, but it is effective only under FedAvg. Our method extends its impact to five more defenses (Multi-Krum, Bulyan, Median, Trim, and FLD), raising the average error rate by $0.399$.
    \item Trim and Krum attack: Our method enables these two attacks to successfully penetrate all baseline defenses (except for Trim attack against FLD) including FLTrust, which were previously unbreachable, with average error rate increases of $0.249$ and $0.253$, respectively.
    \item Min-Max attack: With our method, the Min-Max Attack shows a comprehensive improvement against all defenses except for a slight decrease against Bulyan, achieving an overall average error rate increase of $0.222$.
\end{itemize}
Similarly, the detailed improvements for a specific attack in the non-IID data setting shown as follows:
\begin{itemize}
    \item Sign-flipping attack: Our method helps the sign-flipping attack achieve an average error rate increase of $0.404$, which is similar to the error rate increase on IID data.
    \item Trim and Krum attack: Both attacks penetrate all baseline defenses under the enhancement of our method, with average improvements of $0.281$ and $0.236$, respectively. 
    \item Min-Max attack: Our method helps the Min-Max attack achieve an average error rate increase of $0.195$, higher than its original version. Although this error rate increase is lower than that in the IID data setting, it remains higher than the error rate increase caused by its original version.
\end{itemize}

\begin{table*}[t]
	\centering
        \small
        \setlength{\tabcolsep}{1.8pt}
        \resizebox{0.98\linewidth}{!}{
	    \begin{tabular}{cccccccc|ccccccc}
		\toprule
        \textbf{Data Distribution} & \multicolumn{7}{c}{\textbf{IID}} & \multicolumn{7}{c}{\textbf{Non-IID}}\\
        \otoprule
		\textbf{Attack} & \textbf{FedAvg} & \textbf{FLTrust} & \textbf{MKrum} & \textbf{Bulyan} & \textbf{Median} & \textbf{Trim} & \textbf{FLD} & \textbf{FedAvg} & \textbf{FLTrust} & \textbf{MKrum} & \textbf{Bulyan} & \textbf{Median} & \textbf{Trim} & \textbf{FLD} \\
		\otoprule
		No Attack & 0.028 & 0.051 & 0.029 & 0.029 & 0.045 & 0.029 & 0.025 & 0.029 & 0.042 & 0.030 & 0.029 & 0.041 & 0.029 & 0.022\\
        \midrule
		Sign-Flipping & \textbf{0.934} & 0.059 & 0.038 & 0.055 & 0.055 & 0.036 & 0.025 & \textbf{0.886} & \textbf{0.073} & 0.041 & 0.052 & 0.059 & 0.041 & 0.026\\
		\cmidrule{2-15}
        \rowcolor{BlueGray2}
        \cellcolor{White} + Poison Pill & 0.353 & \textbf{0.093} & \textbf{0.454} & \textbf{0.283} & \textbf{0.268} & \textbf{0.173} & \textbf{0.588} & 0.431 & 0.059 & \textbf{0.605} & \textbf{0.349} & \textbf{0.333} & \textbf{0.217} & \textbf{0.713}\\
		\midrule
		Trim Attack & 0.257 & 0.065 & 0.182 & 0.103 & 0.106 & \textbf{0.123} & 0.022 & 0.418 & 0.059 & 0.295 & 0.209 & 0.245 & \textbf{0.310} & 0.021\\
		\cmidrule{2-15}
        \rowcolor{BlueGray2}
        \cellcolor{White} + Poison Pill & \textbf{0.416} & \textbf{0.109} & \textbf{0.469} &\textbf{0.252} & \textbf{0.247} & 0.117 & \textbf{0.026} & \textbf{0.581} & \textbf{0.065} & \textbf{0.672} & \textbf{0.358} & \textbf{0.324} & \textbf{0.092} & \textbf{0.051} \\
		\midrule
		Krum Attack & 0.033 & 0.061 & 0.067 & 0.154 & 0.188 & 0.043 & \textbf{0.759} & 0.034 & 0.058 & 0.130 & 0.297 & 0.191 & 0.052 & \textbf{0.908}\\
		\cmidrule{2-15}
        \rowcolor{BlueGray2}
        \cellcolor{White} + Poison Pill & \textbf{0.326} & \textbf{0.082} & \textbf{0.585} & \textbf{0.266} & \textbf{0.272} & \textbf{0.169} & 0.632 & \textbf{0.528} & \textbf{0.062} & \textbf{}0.556 & \textbf{0.350} & \textbf{0.321} & \textbf{0.210} & 0.746\\
        \midrule
        Min-Max Attack & 0.307 & 0.082 & \textbf{0.693} & \textbf{0.731} & \textbf{0.341} & \textbf{0.255} & \textbf{0.915} & 0.359 & \textbf{0.161} & \textbf{0.718} & \textbf{0.993} & \textbf{0.381} & \textbf{0.320} & 0.853\\
		\cmidrule{2-15}
        \rowcolor{BlueGray2}
        \cellcolor{White} + Poison Pill & \textbf{0.402} & \textbf{0.106} & 0.518 & 0.273 & 0.262 & 0.218 & 0.766 & \textbf{0.534} & 0.077 & 0.707 & 0.369 & 0.318 & 0.194 & \textbf{0.861}\\
		\bottomrule
	    \end{tabular}
        }
    \caption{Error rates on MNIST under cross-silo setting with 20\% malicious clients in the 50-client FL system.}
	\label{Table:mnist-cs-20}
\end{table*}
\begin{table*}[t]
	\centering
	\small
        \setlength{\tabcolsep}{3.2pt}
	    \begin{tabular}{ccccccc|cccccc}
		\toprule
        \textbf{Data Distribution} & \multicolumn{6}{c}{\textbf{IID}} & \multicolumn{6}{c}{\textbf{Non-IID}}\\
        \otoprule
		\textbf{Attack} & \textbf{FedAvg} & \textbf{FLTrust} & \textbf{MKrum} & \textbf{Bulyan} & \textbf{Median} & \textbf{Trim} & \textbf{FedAvg} & \textbf{FLTrust} & \textbf{MKrum} & \textbf{Bulyan} & \textbf{Median} & \textbf{Trim}\\
		\otoprule
		\multirow{1}{*}{No Attack} & 0.107 &  0.111 &  0.108 &  0.105 &  0.138 &  0.106 & 0.113 &  0.124 &  0.115 &  0.118 &  0.164 &  0.116\\
        \midrule
		\multirow{1}{*}{Sign-Flipping} & \textbf{0.940} &  0.116 &  0.110 &  0.128 &  0.165 &  0.121 & \textbf{0.905} &  0.124 &  0.118 &  0.136 &  0.184 &  0.134\\
		\cmidrule{2-13}
        \rowcolor{BlueGray2}
        \cellcolor{White} + Poison Pill & 0.591 &  \textbf{0.117} &  \textbf{0.749} &  \textbf{0.357} &  \textbf{0.589} &  \textbf{0.225} & 0.573 &  \textbf{0.125} &  \textbf{0.665} &  \textbf{0.379} &  \textbf{0.662} &  \textbf{0.277}\\
		\midrule
		\multirow{1}{*}{Trim Attack} & 0.240 &  0.110 &  0.151 &  0.148 &  0.207 &  0.178 & 0.340 &  0.120 &  0.228 &  0.190 &  0.237 &  \textbf{0.245}\\
		\cmidrule{2-13}
        \rowcolor{BlueGray2}
        \cellcolor{White} + Poison Pill & \textbf{0.620} &  \textbf{0.492} &  \textbf{0.620} &  \textbf{0.228} &  \textbf{0.424} &  \textbf{0.232} & \textbf{0.654} &  \textbf{0.533} &  \textbf{0.679} &  \textbf{0.324} &  \textbf{0.483} &  0.226\\
		\midrule
		\multirow{1}{*}{Krum Attack} & 0.117 &  0.112 &  0.172 &  0.238 &  0.169 &  0.132 & 0.126 &  0.121 &  0.204 &  0.296 &  0.222 &  0.158\\
		\cmidrule{2-13}
        \rowcolor{BlueGray2}
        \cellcolor{White} + Poison Pill & \textbf{0.681} &  \textbf{0.138} &  \textbf{0.740} &  \textbf{0.362} &  \textbf{0.572} &  \textbf{0.258} & \textbf{0.604} &  \textbf{0.141} &  \textbf{0.750} &  \textbf{0.372} &  \textbf{0.649} &  \textbf{0.277}\\
        \midrule
        \multirow{1}{*}{Min-Max Attack} & 0.146 &  0.111 &  0.382 &  \textbf{0.324} &  0.183 &  0.185 & 0.191 &  0.147 &  0.621 &  \textbf{0.426} &  0.245 &  0.279\\
		\cmidrule{2-13}
        \rowcolor{BlueGray2}
        \cellcolor{White} + Poison Pill & \textbf{0.651} &  \textbf{0.244} &  \textbf{0.718} &  0.312 &  \textbf{0.503} &  \textbf{0.249} & \textbf{0.670} &  \textbf{0.229} &  \textbf{0.621} &  0.349 &  \textbf{0.581} &  \textbf{0.386}\\
		\bottomrule
	    \end{tabular}
        \caption{Error rates on Fashion-MNIST under cross-device setting with 20\% malicious clients in 50-client FL system.}
	\label{Table:fmnist-cd-iid}
\end{table*}
\begin{table*}[t]
	\centering
	\small
        \setlength{\tabcolsep}{3.2pt}
	    \begin{tabular}{ccccccc|cccccc}
		\toprule
        \textbf{Data Distribution} & \multicolumn{6}{c}{\textbf{IID}} & \multicolumn{6}{c}{\textbf{Non-IID}}\\
        \otoprule
		\textbf{Attack} & \textbf{FedAvg} & \textbf{FLTrust} & \textbf{MKrum} & \textbf{Bulyan} & \textbf{Median} & \textbf{Trim} & \textbf{FedAvg} & \textbf{FLTrust} & \textbf{MKrum} & \textbf{Bulyan} & \textbf{Median} & \textbf{Trim}\\
		\otoprule
		\multirow{1}{*}{No Attack} & 0.110 &  0.106 &  0.107 &  0.108 &  0.139 &  0.106 & 0.115 &  0.117 &  0.115 &  0.117 &  0.164 &  0.112\\
        \midrule
		\multirow{1}{*}{Sign-Flipping} & \textbf{0.929} &  0.111 &  0.108 &  0.111 &  0.153 &  0.117 & \textbf{0.902} &  0.118 &  0.115 &  0.120 &  0.175 &  0.134\\
		\cmidrule{2-13}
        \rowcolor{BlueGray2}
        \cellcolor{White} + Poison Pill & 0.195 &  \textbf{0.114} &  \textbf{0.170} &  \textbf{0.138} &  \textbf{0.347} &  \textbf{0.137} & 0.330 &  \textbf{0.124} &  \textbf{0.165} &  \textbf{0.148} &  \textbf{0.483} &  \textbf{0.161}\\
		\midrule
		\multirow{1}{*}{Trim Attack} & 0.112 &  0.114 &  0.111 &  0.118 &  0.153 &  0.113 & 0.129 &  0.125 &  0.128 &  0.129 &  0.185 &  0.122\\
		\cmidrule{2-13}
        \rowcolor{BlueGray2}
        \cellcolor{White} + Poison Pill & \textbf{0.369} &  \textbf{0.138} &  \textbf{0.212} &  \textbf{0.128} &  \textbf{0.310} &  \textbf{0.140} & \textbf{0.589} &  \textbf{0.154} &  \textbf{0.300} &  \textbf{0.139} &  \textbf{0.351} &  \textbf{0.156}\\
		\midrule
		\multirow{1}{*}{Krum Attack} & 0.110 &  0.113 &  0.115 &  0.128 &  0.144 &  0.113 & 0.121 &  0.116 &  0.123 &  0.135 &  0.183 &  0.120\\
		\cmidrule{2-13}
        \rowcolor{BlueGray2}
        \cellcolor{White} + Poison Pill & \textbf{0.164} &  \textbf{0.117} &  \textbf{0.157} &  \textbf{0.143} &  \textbf{0.371} &  \textbf{0.142} & \textbf{0.229} &  \textbf{0.126} &  \textbf{0.249} &  \textbf{0.146} &  \textbf{0.374} &  \textbf{0.157}\\
        \midrule
        \multirow{1}{*}{Min-Max Attack} & 0.116 &  0.111 &  0.116 &  0.127 &  0.145 &  0.122 & 0.121 &  0.116 &  0.123 &  0.135 &  0.183 &  0.120\\
		\cmidrule{2-13}
        \rowcolor{BlueGray2}
        \cellcolor{White} + Poison Pill & \textbf{0.351} &  \textbf{0.124} &  \textbf{0.299} &  \textbf{0.135} &  \textbf{0.343} &  \textbf{0.146} & \textbf{0.342} &  \textbf{0.138} &  \textbf{0.292} &  \textbf{0.148} &  \textbf{0.417} &  \textbf{0.166}\\
		\bottomrule
	    \end{tabular}
        \caption{Error rates on Fashion-MNIST under cross-device setting with 10\% malicious clients in 50-client FL system.}
	\label{Table:fmnist-cd-10}
\end{table*}
\section{Additional Stealthiness Analysis}
\label{appendix:stealthy}

\begin{figure}[t]
    \centering
    \includegraphics[width = 1\linewidth]{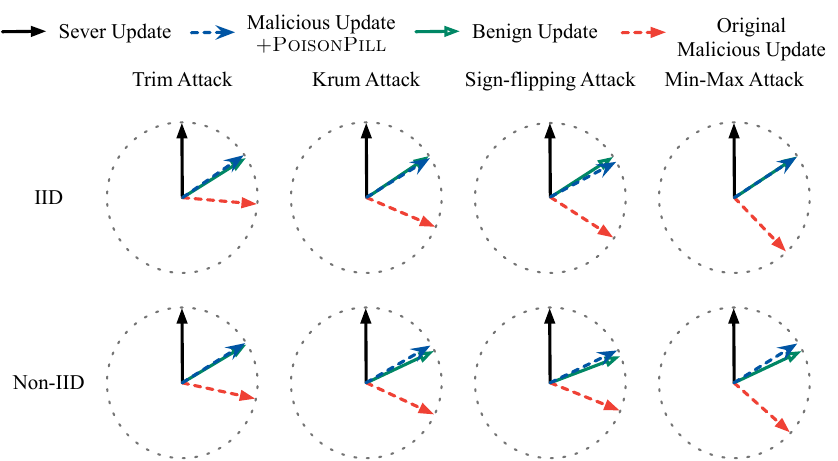}
    \caption{Comparison of cosine similarity scores between original attack with and without our method under FLTrust.}
    \label{fig:iid similarity}
\end{figure}
\begin{figure*}[t]
    \centering
    \includegraphics[width = 0.98\linewidth]{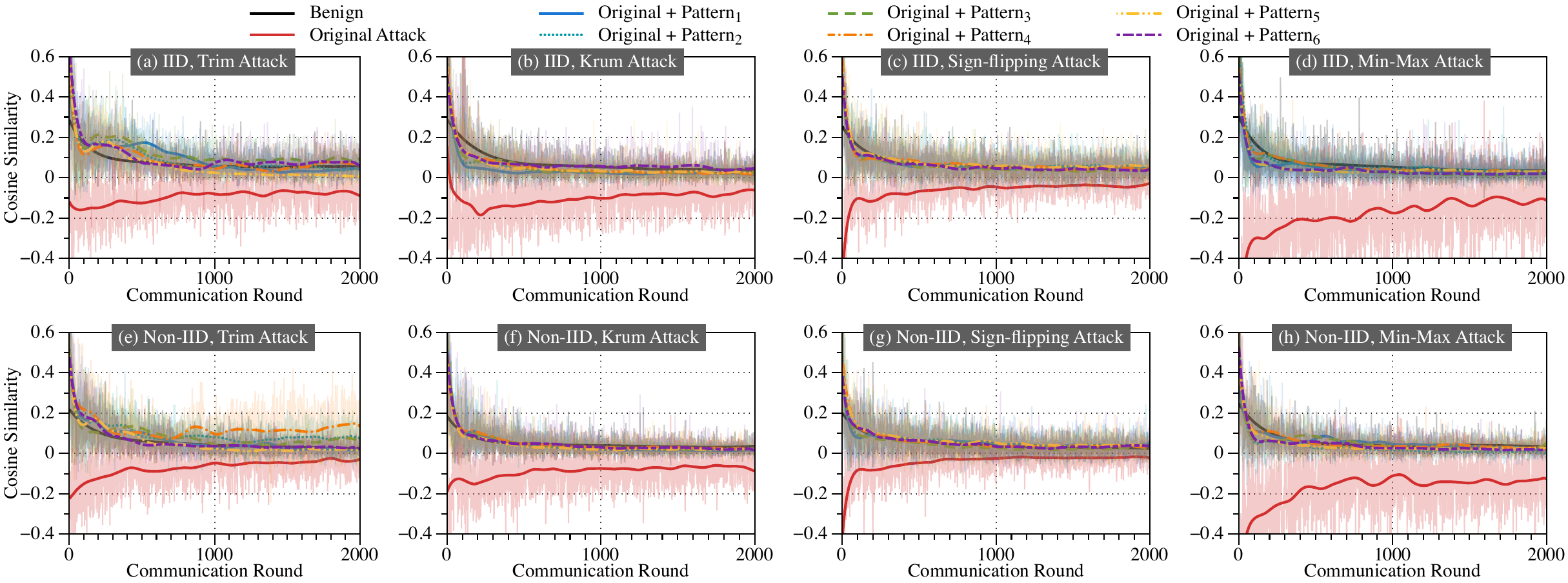}
    \caption{Comparison of the cosine similarity scores with the server model of the original attack and our different augmentation patterns in the entire training period under FLTrust.}
    \label{fig:complete similarity}
\end{figure*}

\parab{\ul{Distance Score Analysis.}}
Figure~\ref{fig:iid distance} compares the average distance scores of benign and malicious clients (with and without our method) across four baseline model poisoning attacks. The distance scores when using our method closely match or are even identical to those of benign clients throughout the entire training process. In contrast, original attacks like the Trim and sign-flipping attacks display distance scores that were significantly higher or lower than those of benign updates, indicating either detected by Multi-Krum (higher scores) or underutilized attack capacities (lower scores). Our method also has a lower distance score variance in the early FL training period, representing that our method provides more steady attack efficacy in the FL's critical training period~\citep{yan2023defl,yan2022seizing} by fully utilizing the attack capacities while being undetected. Additionally, our method also achieves two more improvements. First, our method causes the global model to degrade earlier compared to the original attacks, further demonstrating the effectiveness of our augmentation. Second, our method significantly increases the discrepancy between benign client updates as the communication rounds increase. While original attacks can bypass detection in some cases, the discrepancy between benign client updates remains steady, illustrating the lower impact of malicious clients. In contrast, our method consistently increases the discrepancy among benign clients, highlighting its penetrating effectiveness in its influence on benign clients' local training.

\parab{\ul{Cosine Similarity Score Analysis.}}
Figure~\ref{fig:iid similarity} shows that the angles between server model updates and malicious updates using our method are similar or even smaller than those of benign updates, leading to higher aggregation weights for malicious updates in FLTrust -- illustrating why our method makes existing FL poisoning attacks effectively bypass FLTrust. In contrast, the angles between the FLTrust's server model updates with original malicious updates are often greater than $90^{\circ}$, leading to a zero aggregation weight. Detailed per-round cosine similarity trends (Figure~\ref{fig:complete similarity}) also reveal that while original attacks often result in negative similarities (and thus are excluded by FLTrust), our method maintains positive similarities throughout the entire training process. This consistency not only ensures the successful insertion of the pill in any specific round but also secures pill's long-lasting presence in the global model.

\section{Additional Details on MNIST and CIFAR-10 Dataset}
\label{appendix:mnist cifar}
The detailed results on MNIST and CIFAR-10 datasets are presented in Table~\ref{Table:mnist-cs-20} (MNIST dataset) and Table~\ref{Table:cifar-cs-20} (CIFAR-10 dataset), respectively.

For MNIST dataset, the highest error rate increase achieved using our method is $0.518$, with an average increase of $0.121$. This average error rate increase is slightly lower compared with the improvement observed on the Fashion-MNIST dataset. Despite the reduced average error rate increase, it remains significant, especially considering the MNIST dataset's lower baseline error rates (below 0.070). 

On CIFAR-10 dataset, our method helps existing FL poisoning attacks outperform their original versions in $71$ of the $72$ scenarios, with an average error rate increase of over $0.288$. Specifically, our method facilitates at least a $0.212$ increase in error rates against FLTrust, outperforming the results in the same settings on the Fashion-MNIST dataset.

\section{Additional Results in Cross-device FL System}
\label{appendix:cd}

After evaluating our method in the 50-client cross-silo FL system, we further test it in the 50-client cross-device FL system. Table~\ref{Table:fmnist-cd-iid} presents the error rates under the cross-device FL setting using the ``approximate max pill search'' algorithm on both IID and non-IID data. We report the highest error rates among the results of six dynamic patterns, with the malicious client proportion set to $20\%$. Since FLD is not typically designed for cross-device systems, we do not test it in this setting.

\parab{\ul{Results on IID Data.}} The highest error rate improvement with our method achieves $0.639$, and the average error rate increase with our method reaches $0.279$. With our method, existing model poisoning attacks outperform their original versions in $22$ out of the $24$ cases. The highest error rate improvement for the sign-flipping attack is $0.639$, with an average error rate increase of $0.279$. For the Trim attack and Krum attack, the highest error rate increases are $0.469$ and $0.568$, with average error rate increases of $0.264$ and $0.302$, respectively. For the Min-Max attack, the highest error rate increase reaches $0.505$, with an average increase of $0.272$. These improvements are consistent with the error rates observed under the cross-silo FL setting using the "approximate max pill search" algorithm on IID data.

\parab{\ul{Results on non-IID Data.}} As for the results on non-IID data, the highest error rate improvement with our method achieves $0.546$, and the average error rate increase with our method reaches $0.273$. By using our method, existing model poisoning attacks outperform their original versions in 21 out of the 24 cases. The highest error rate improvement for the sign-flipping attack is $0.547$, with an average error rate increase of $0.282$. For the Trim attack and Krum attack, the highest error rate rises are $0.451$ and $0.546$, with an average error rate rise of $0.312$ and $0.278$, respectively. For the Min-Max attack, the highest error rate increase reaches $0.479$, with an average increase of $0.201$. These improvements are also aligned with the error rates observed under the cross-silo FL setting using the max subnetwork searching algorithm on non-IID data.

The average error rates of the global model in the cross-device FL system are lower than the error rates in the cross-silo FL system within $0.030$, illustrating our method's generality over different data distribution and FL systems.

\section{Additional Results with Fewer Malicious Clients}
\label{appendix:10client}
We also test the error rate improvement of our method in both the IID and non-IID cross-device FL systems, with only $10\%$ malicious clients. The experimental results are shown in Table~\ref{Table:fmnist-cd-10}. 

\parab{\ul{Results on IID Data with Fewer Malicious Clients.}} The highest error rate increment is $0.257$, with an average increment of $0.083$. The error rate increments in the cross-device FL system are smaller than those in the cross-silo FL system, as malicious clients may not be selected in every round. However, this reduction in improvement is acceptable since our method helps existing model poisoning attacks outperform their original versions in $23$ out of $24$ cases. Furthermore, when all existing attacks fail to bypass any defenses with $10\%$ malicious clients, our method enables the attacks to bypass all defenses. The superiority of our method is maintained even with $10\%$ compromised clients.

\parab{\ul{Results on Non-IID Data with Fewer Malicious Clients.}} The results on the non-IID data are similar to those on the IID data. The highest error rate increment is $0.460$, and the average error rate increment is $0.079$. Our method helps existing model poisoning attacks achieve higher error rates in $23$ out of $24$ cases, even in highly unstable and heterogeneous settings. These results demonstrate the generality and robustness of our method across different data distributions and client selection methods with only a small portion of malicious clients.

\section{Impact of The Pill Search Algorithm}
\label{appendix:impact search}
\begin{figure*}[ht]
    \centering
    \includegraphics[width = 0.98\textwidth]{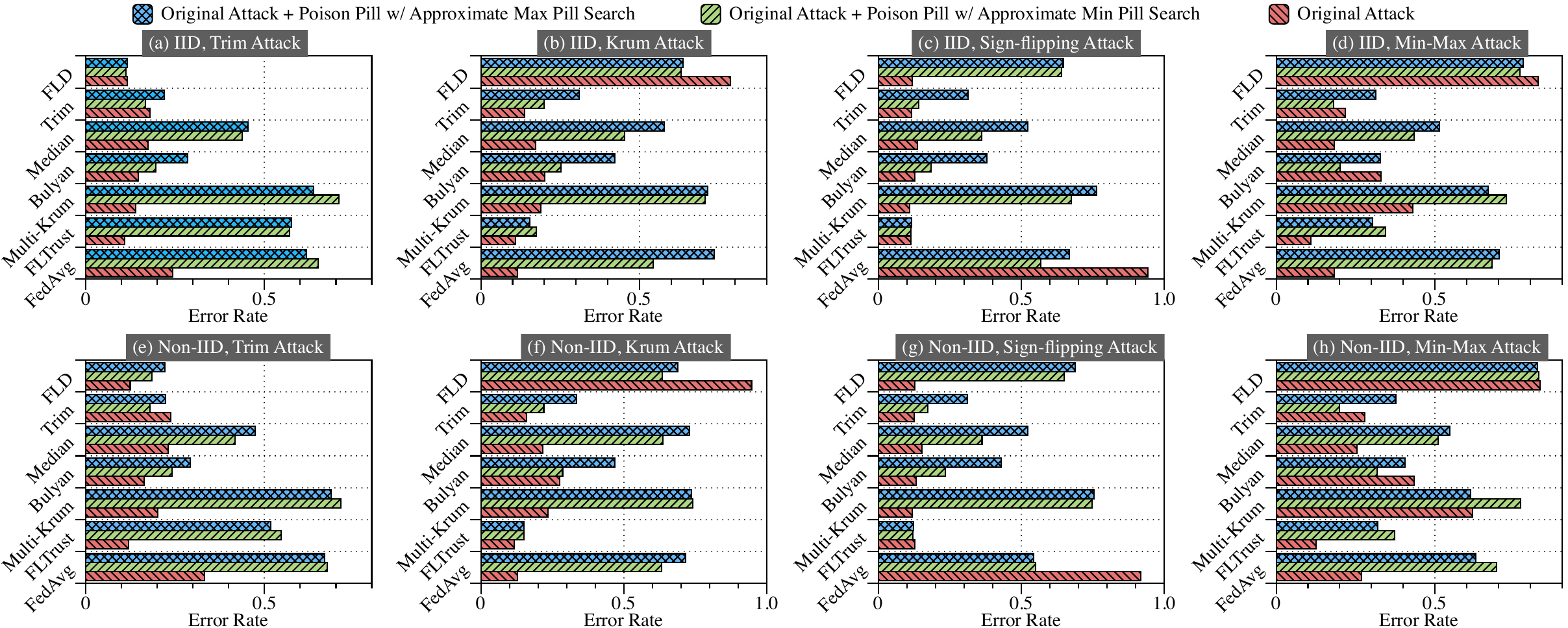}
    \caption{Comparison of error rates between original poisoning attacks and attacks enhanced by our method using two different pill search methods.}
    \label{fig:comparison subnetwork selection}
\end{figure*}
We conduct a final evaluation to assess the importance and effectiveness of the "approximate max pill search" algorithm used in our method. This is contrasted against a newly devised "approximate min pill search" algorithm, which targets the least important parameters within the target model. Figure~\ref{fig:comparison subnetwork selection} illustrates the error rates achieved by the "approximate max pill search", the "approximate min pill search", and the original model poisoning attacks. The "approximate max pill search" algorithm outperforms the "approximate min pill search" in $41$ out of $56$ cases (approximately $73$\%), underscoring its effectiveness in leveraging the most influential parameters to enhance attack impacts. Despite its lower efficacy, the "approximate min pill search" still manages to surpass the original attacks in $41$ out of $56$ cases (approximately $73$\%). This demonstrates the generality of our method across different pill search algorithms.

\section{Our Method Against Adaptive Defense}
\label{appendix:adaptive defense}
We develop an adaptive defense named DSTrust, which enhances the FLTrust's mechanism. DSTrust incorporates both distance and cosine similarity scores into a unified trust score calculation, directly countering our method’s two-step adjustment approach. The round-$t$ trust score of client $i$ in DSTrust is calculated as follows:
\begin{equation}
    TS_i = ReLU(\frac{\mathop{cos}(\Delta\boldsymbol{g}_t^{(i)}, \Delta\boldsymbol{g}_t^s)}{||\Delta\boldsymbol{g}_t^{(i)}-\Delta\boldsymbol{g}_t^s||}),
\end{equation}
where $\Delta\boldsymbol{g}_t^{(i)}$ represents the model update from client $i$ and $\Delta\boldsymbol{g}_t^s)$ represents server's model update. By integrating both cosine similarity and distance metrics, DSTrust provides a more comprehensive defense approach compared with FLTrust. This dual consideration allows DSTrust to effectively mitigate attacks that manipulate either of these metrics to bypass defenses.

\begin{table}[t]
	\centering
        \small
        \setlength{\tabcolsep}{0.8pt}
	    \begin{tabular}{ccc|cc}
		\toprule
        \textbf{Distribution} & \multicolumn{2}{c}{\textbf{IID}} & \multicolumn{2}{c}{\textbf{Non-IID}}\\
        \otoprule
		\textbf{Attack} & \Centerstack{w/o \\\ Poison Pill} &\Centerstack{w/ \\\ Poison Pill} & \Centerstack{w/o \\\ Poison Pill} & \Centerstack{w/ \\\ Poison Pill} \\
		\otoprule
		No Attack &  \multicolumn{2}{c|}{0.108}  &  \multicolumn{2}{c}{0.116}  \\
        \midrule
		Sign-Flipping & 0.111 & \cellcolor{BlueGray2}\textbf{0.129} & 0.110 & \cellcolor{BlueGray2}\textbf{0.131} \\
		\midrule
		Trim Attack & 0.109 & \cellcolor{BlueGray2}\textbf{0.629} & 0.115 & \cellcolor{BlueGray2}\textbf{0.630} \\
		\midrule
		Krum Attack & 0.111 & \cellcolor{BlueGray2}\textbf{0.140} & 0.120 & \cellcolor{BlueGray2}\textbf{0.128} \\
        \midrule
        Min-Max Attack & 0.127 & \cellcolor{BlueGray2}\textbf{0.167} & 0.143 & \cellcolor{BlueGray2}\textbf{0.327} \\
		\bottomrule
	    \end{tabular}
    \caption{Error rates under cross-silo setting against the new adaptive defense -- DSTrust -- with and without our method on Fashion-MNIST dataset (20\% malicious clients).}
	\label{Table:adaptive}
\end{table}
\begin{table*}[htbp]
	\centering
    \small
        \setlength{\tabcolsep}{1.8pt}
	    \begin{tabular}{cccccccc|ccccccc}
		\toprule
        \textbf{Data Distribution} & \multicolumn{7}{c}{\textbf{IID}} & \multicolumn{7}{c}{\textbf{Non-IID}}\\
        \otoprule
		\textbf{Attack} & \textbf{FedAvg} & \textbf{FLTrust} & \textbf{MKrum} & \textbf{Bulyan} & \textbf{Median} & \textbf{Trim} & \textbf{FLD} & \textbf{FedAvg} & \textbf{FLTrust} & \textbf{MKrum} & \textbf{Bulyan} & \textbf{Median} & \textbf{Trim} & \textbf{FLD}\\
		\otoprule
        Sign-Flipping & \textbf{0.943} & 0.114 & 0.108 & 0.126 & 0.136 & 0.116 & 0.118  & \textbf{0.917} & \textbf{0.126} & 0.117 & 0.132 & 0.152 & 0.124 & 0.127 \\
        \cmidrule{2-15}
        \Centerstack{+ Poison Pill \\w/o SimAdjust \\w/ DistAdjust} & 0.311 & 0.096 & 0.219 & \textbf{0.413} & 0.354 & 0.381 & \textbf{0.726} & 0.372 & 0.111 & 0.224 & 0.373 & 0.298 & 0.448 & \textbf{0.722} \\
        \cmidrule{2-15}
        \Centerstack{+ Poison Pill \\w/ SimAdjust \\w/o DistAdjust} & 0.826 & 0.110 & 0.118 & 0.142 & \textbf{0.898} & \textbf{0.464} & 0.695 & 0.849 & 0.117 & 0.133 & 0.161 & \textbf{0.870} & \textbf{0.491} & 0.634 \\
        \cmidrule{2-15}
        \rowcolor{BlueGray2}
        \Centerstack{\cellcolor{White}+ Poison Pill \\w/ SimAdjust \\w/ DistAdjust} & 0.667 & \textbf{0.115} & \textbf{0.764} & 0.379 & 0.523 & 0.314 & 0.646 & 0.543 & 0.122 & \textbf{0.754} & \textbf{0.430} & 0.522 & 0.311 & 0.688 \\
        \midrule
        Trim Attack & 0.243 & 0.109 & 0.139 & 0.146 & 0.174 & 0.179 & 0.116 & 0.332 & 0.120 & 0.201 & 0.163 & 0.231 & 0.238 & 0.124\\
        \cmidrule{2-15}
        \Centerstack{+ Poison Pill \\w/o SimAdjust \\w/ DistAdjust} & 0.317 & 0.105 & 0.364 & 0.247 & 0.368 & 0.136 & \textbf{0.208} & 0.219 & 0.111 & 0.600 & \textbf{0.456} & 0.295 & 0.212 & 0.105 \\
        \cmidrule{2-15}
        \Centerstack{+ Poison Pill \\w/ SimAdjust \\w/o DistAdjust} & 0.554 & 0.104 & 0.122 & 0.108 & 0.429 & \textbf{0.284} & 0.119 & \textbf{0.744} & 0.110 & 0.134 & 0.126 & \textbf{0.898} & \textbf{0.478} & 0.109 \\
        \cmidrule{2-15}
        \rowcolor{BlueGray2}
        \Centerstack{\cellcolor{White}+ Poison Pill \\w/ SimAdjust \\w/ DistAdjust} & \textbf{0.618} & \textbf{0.576} & \textbf{0.638} & \textbf{0.284} & \textbf{0.453} & 0.219 & 0.115 & 0.668 & \textbf{0.517} & \textbf{0.687} & 0.292 & 0.473 & 0.223 & \textbf{0.222} \\
		\bottomrule
	    \end{tabular}
    \caption{Ablation study on the necessity of both SimAdjust and DistAdjust in our method on Fashion-MNIST dataset with 20\% malicious clients in the 50-client FL system.}
	\label{Table:ablation adjustment}
\end{table*}
\begin{table*}[t]
	\centering
    \small
        \setlength{\tabcolsep}{1.8pt}
	    \begin{tabular}{cccccccc|ccccccc}
		\toprule
        \textbf{Data Distribution} & \multicolumn{7}{c}{\textbf{IID}} & \multicolumn{7}{c}{\textbf{Non-IID}}\\
        \otoprule
		\textbf{Attack} & \textbf{FedAvg} & \textbf{FLTrust} & \textbf{MKrum} & \textbf{Bulyan} & \textbf{Median} & \textbf{Trim} & \textbf{FLD} & \textbf{FedAvg} & \textbf{FLTrust} & \textbf{MKrum} & \textbf{Bulyan} & \textbf{Median} & \textbf{Trim} & \textbf{FLD}\\
		\otoprule
		No Attack & 0.109 & 0.107 & 0.105 & 0.105 & 0.123 & 0.106 & 0.115 & 0.113 & 0.115 & 0.115 & 0.112 & 0.142 & 0.115 & 0.122 \\
        \midrule
		Label-Flipping & \textbf{0.937} & 0.110 & 0.103 & 0.115 & 0.125 & 0.122 & 0.097 & \textbf{0.767} & 0.118 & 0.108 & 0.128 & 0.126 & 0.125 & 0.104 \\
		\cmidrule{2-15} 
        \rowcolor{BlueGray2}
        \cellcolor{White} + Poison Pill & 0.464 & \textbf{0.124} & \textbf{0.592} & \textbf{0.466} & \textbf{0.897} & \textbf{0.455} & \textbf{0.917} & 0.471 & \textbf{0.124} & \textbf{0.769} & \textbf{0.646} & \textbf{0.902} & \textbf{0.883} & \textbf{0.927} \\
		\bottomrule
	    \end{tabular}
        \caption{Error rates of Label-flipping Attack on Fashion-MNIST under cross-silo setting with 20\% malicious clients in the 50-client FL system.}
	\label{Table:label flip}
\end{table*}

Table~\ref{Table:adaptive} details the error rates for four baseline FL poisoning attacks both with and without our method against the DSTrust defense on the Fashion-MNIST dataset within a 50-client FL system, where 20\% of clients are malicious. These tests were conducted under both IID and non-IID data environments. DSTrust effectively neutralizes the four baseline poisoning attacks when our method is not applied, highlighting its robustness as a defense mechanism. Despite DSTrust's integration of both cosine similarity and distance metrics in its defense strategy, it fails to counteract the augmented attacks when our method is employed. Notably, our method achieves a maximum error rate increase of $0.521$, and an average error rate increase of $0.173$ across all $8$ test scenarios. These results demonstrate that merely understanding the adjustment strategies of our method, and subsequently integrating corresponding defense metrics, does not fundamentally negate the effectiveness of our method. Despite the adaptive defense's attempt to incorporate both cosine similarity and distance metrics into DSTrust, it remains insufficient to thwart the enhanced capabilities of our method.

\begin{table*}[t]
	\centering
    \small
        \setlength{\tabcolsep}{1.2pt}
	    \begin{tabular}{cccccccccc|ccccccccc}
		\toprule
        \textbf{Distribution} & \multicolumn{9}{c}{\textbf{IID}} & \multicolumn{9}{c}{\textbf{Non-IID}}\\
        \otoprule
		\textbf{Attack} & \textbf{FAvg} & \textbf{FLT} & \textbf{MKr} & \textbf{Bulyan} & \textbf{Median} & \textbf{Trim} & \textbf{DnC} & \textbf{FLD} & \textbf{Flame} & \textbf{FAvg} & \textbf{FLT} & \textbf{MKr} & \textbf{Bulyan} & \textbf{Median} & \textbf{Trim} & \textbf{DnC} & \textbf{FLD} & \textbf{Flame}\\
		\otoprule
		No Attack & 0.319 & 0.328 & 0.338 & 0.324 & 0.330 & 0.337 & 0.315 & 0.334 & 0.336 & 0.336 & 0.358 & 0.330 & 0.336 & 0.338 & 0.336 & 0.347 & 0.326 & 0.374 \\
        \midrule
		Sign-Flipping & \textbf{0.897} & 0.335 & 0.336 & 0.329 & 0.353 & 0.386 & 0.341 & 0.316 & 0.367 & \textbf{0.894} & 0.301 & 0.361 & 0.317 & 0.365 & 0.433 & 0.317 & 0.315 & 0.378 \\
		\cmidrule{2-19}
        \rowcolor{BlueGray2}
        \cellcolor{White} + Poison Pill & 0.711 & \textbf{0.483} & \textbf{0.503} & \textbf{0.457} & \textbf{0.385} & \textbf{0.410} & \textbf{0.413} & \textbf{0.898} & \textbf{0.487} & 0.471 & \textbf{0.551} & \textbf{0.593} & \textbf{0.482} & \textbf{0.459} & \textbf{0.448} & \textbf{0.415} & \textbf{0.731} & \textbf{0.517} \\
        \midrule
		Trim Attack & 0.431 & 0.323 & 0.422 & 0.428 & \textbf{0.434} & \textbf{0.432} & 0.340 & \textbf{0.339} & 0.362 & 0.423 & 0.345 & 0.469 & \textbf{0.429} & 0.474 & \textbf{0.451} & 0.352 & \textbf{0.299} & 0.355 \\
		\cmidrule{2-19}
        \rowcolor{BlueGray2}
        \cellcolor{White} + Poison Pill & \textbf{0.490} & \textbf{0.578} & \textbf{0.595} & \textbf{0.506} & 0.428 & 0.392 & \textbf{0.406} & 0.295 & \textbf{0.383} & \textbf{0.578} & \textbf{0.478} & \textbf{0.540} & 0.407 & \textbf{0.823} & 0.420 & \textbf{0.396} & 0.283 & \textbf{0.378} \\
        \midrule
		Krum Attack & 0.324 & 0.374 & 0.441 & \textbf{0.523} & \textbf{0.471} & 0.393 & 0.348 & 0.609 & 0.360 & 0.379 & 0.300 & 0.479 & \textbf{0.547} & \textbf{0.492} & \textbf{0.397} & 0.330 & 0.704 & 0.352 \\
		\cmidrule{2-19}
        \rowcolor{BlueGray2}
        \cellcolor{White} + Poison Pill & \textbf{0.393} & \textbf{0.458} & \textbf{0.528} & 0.480 & 0.426 & \textbf{0.413} & \textbf{0.426} & \textbf{0.908} & \textbf{0.550} & \textbf{0.502} & \textbf{0.557} & \textbf{0.578} & 0.458 & 0.414 & 0.380 & \textbf{0.399} & \textbf{0.924} & \textbf{0.535} \\
		\bottomrule
	    \end{tabular}
    \caption{Error rates on CIFAR-10 under cross-silo setting with 20\% malicious clients and VGG-11 Net in the 30-client FL system.}
	\label{Table:vgg}
\end{table*}
\begin{table*}[t]
	\centering
    \small
        \setlength{\tabcolsep}{1.8pt}
	    \begin{tabular}{cccccccc|ccccccc}
		\toprule
        \textbf{Data Distribution} & \multicolumn{7}{c}{\textbf{IID}} & \multicolumn{7}{c}{\textbf{Non-IID}}\\
        \otoprule
		\textbf{Attack} & \textbf{FedAvg} & \textbf{FLTrust} & \textbf{MKrum} & \textbf{Bulyan} & \textbf{Median} & \textbf{Trim} & \textbf{FLD} & \textbf{FedAvg} & \textbf{FLTrust} & \textbf{MKrum} & \textbf{Bulyan} & \textbf{Median} & \textbf{Trim} & \textbf{FLD}\\
		\otoprule
		No Attack & 0.093 & 0.097 & 0.093 & 0.094 & 0.111 & 0.092 & 0.093  & 0.097 & 0.102 & 0.101 & 0.105 & 0.112 & 0.100 & 0.099 \\
        \midrule
        Sign-Flipping & \textbf{0.893} & 0.157 & 0.092 & 0.106 & 0.110 & 0.108 & 0.100 & \textbf{0.963} & 0.122 & 0.104 & 0.103 & 0.118 & 0.114 & 0.106 \\
		\cmidrule{2-15}
        \rowcolor{BlueGray2}
        \cellcolor{White} + Poison Pill & 0.179 & \textbf{0.167} & \textbf{0.254} & \textbf{0.457} & \textbf{0.463} & \textbf{0.289} & \textbf{0.594} & 0.239 & \textbf{0.165} & \textbf{0.328} & \textbf{0.451} & \textbf{0.604} & \textbf{0.372} & \textbf{0.685} \\
        \midrule
		Trim Attack & 0.274 & \textbf{0.126} & 0.108 & 0.105 & 0.191 & \textbf{0.219} & 0.097  & 0.357 & \textbf{0.135} & 0.134 & \textbf{0.308} & 0.225 & \textbf{0.270} & 0.101 \\
		\cmidrule{2-15}
        \rowcolor{BlueGray2}
        \cellcolor{White} + Poison Pill & \textbf{0.336} & 0.101 & \textbf{0.901} & \textbf{0.281} & \textbf{0.272} & 0.122 & \textbf{0.518}  & \textbf{0.363} & 0.107 & \textbf{0.248} & 0.285 & \textbf{0.305} & 0.138 & \textbf{0.182} \\
		\bottomrule
	    \end{tabular}
    \caption{Error rates on Fashion-MNIST under cross-silo setting with 20\% malicious clients in a 100-client FL system.}
	\label{Table:100-trim}
\end{table*}
\begin{table*}[ht]
	\centering
    \small
        \setlength{\tabcolsep}{1.8pt}
	    \begin{tabular}{cccccccc|ccccccc}
		\toprule
        \textbf{Data Distribution} & \multicolumn{7}{c}{\textbf{IID}} & \multicolumn{7}{c}{\textbf{Non-IID}}\\
        \otoprule
		\textbf{Attack} & \textbf{FedAvg} & \textbf{FLTrust} & \textbf{MKrum} & \textbf{Bulyan} & \textbf{Median} & \textbf{Trim} & \textbf{FLD} & \textbf{FedAvg} & \textbf{FLTrust} & \textbf{MKrum} & \textbf{Bulyan} & \textbf{Median} & \textbf{Trim} & \textbf{FLD}\\
		\otoprule
		No Attack & 0.109 & 0.107 & 0.105 & 0.105 & 0.123 & 0.106 & 0.115 & 0.113 & 0.115 & 0.115 & 0.112 & 0.142 & 0.115 & 0.122 \\
        \midrule
		Sign-Flipping & \textbf{0.943} & 0.114 & 0.108 & 0.126 & 0.136 & 0.116 & 0.118  & \textbf{0.917} & 0.126 & 0.117 & 0.132 & 0.152 & 0.124 & 0.127 \\
        \cmidrule{2-15}
        + Neurotoxin & 0.710 & \textbf{0.147} & 0.105 & 0.103 & 0.106 & 0.105 & 0.110 & 0.739 & \textbf{0.146} & 0.121 & 0.122 & 0.124 & 0.124 & 0.119 \\
		\cmidrule{2-15}
        \rowcolor{BlueGray2}
        \cellcolor{White} + Poison Pill & 0.667 & 0.115 & \textbf{0.764} & \textbf{0.379} & \textbf{0.523} & \textbf{0.314} & \textbf{0.646} & 0.543 & 0.122 & \textbf{0.754} & \textbf{0.430} & \textbf{0.522} & \textbf{0.311} & \textbf{0.688} \\
        \midrule
        Trim Attack & 0.243 & 0.109 & 0.139 & 0.146 & 0.174 & 0.179 & \textbf{0.116} & 0.332 & 0.120 & 0.201 & 0.163 & 0.231 & \textbf{0.238} & 0.124\\
        \cmidrule{2-15}
        + Neurotoxin & 0.135 & 0.109 & 0.113 & 0.106 & 0.126 & 0.119 & 0.108 & 0.168 & 0.187 & 0.144 & 0.197 & 0.138 & 0.144 & 0.121 \\
		\cmidrule{2-15}
        \rowcolor{BlueGray2}
        \cellcolor{White} + Poison Pill & \textbf{0.618} & \textbf{0.576} & \textbf{0.638} & \textbf{0.284} & \textbf{0.453} & \textbf{0.219} & 0.115 & \textbf{0.668} & \textbf{0.517} & \textbf{0.687} & \textbf{0.292} & \textbf{0.473} & 0.223 & \textbf{0.222} \\
		\bottomrule
	    \end{tabular}
        \caption{Comparison with Neurotoxin under cross-silo setting on IID Fashion-MNIST in the 50-client FL system.}
	\label{Table:neurotoxin}
\end{table*}

\section{Limitations and Future Work}
\label{appendix:limitations}
Our method significantly enhances non-state-of-the-art (non-SOTA) model poisoning attacks, enabling them to SOTA results against various prevalent defenses. This is accomplished through a pill-based, attack-agnostic augmentation pipeline. We not only demonstrate our method’s capabilities but also expose fundamental vulnerabilities within the current designs of defense mechanisms.

For future attacks in FL, it is essential for attackers to meticulously evaluate the importance of each parameter in their implementation. By targeting specific subsets of parameters, attackers can devise more flexible and adaptive attacks, improving \textit{stealthiness} and complicating defense efforts. As for future defenses, while individually checking each parameter might seem viable, its practical deployment is hindered by high overheads, making it infeasible in real-world applications.

Thus, there is a pressing need for more sophisticated defenses that can conduct fine-grained analyses of the roles of different parameters in neural networks, while executing without imposing prohibitive computational costs.

\section{Additional Ablation Study on Pill Adjustment}
\label{appendix:ablation adjust}
To illustrate the necessity of both the \FuncCall{SimAdjust}{} and \FuncCall{DistAdjust}{} used in our method, we conduct a detailed ablation study, providing the error rates of the Sign-flipping Attack and the Trim Attack with different settings of pill adjustments on the IID and non-IID Fashion-MNIST dataset within a 50-client FL system in Table~\ref{Table:ablation adjustment}.

The results show that using both \FuncCall{SimAdjust}{} and \FuncCall{DistAdjust}{} together achieves the highest error rate in multiple cases. For Sign-flipping Attack, the combined adjustment method outperforms others against multi-Krum (both IID and non-IID), FLDetector (both IID and non-IID), and Bulyan (non-IID). Similarly, for Trim Attack, the combined approach yields the highest error rates in 8 out of 14 defenses under both IID and Non-IID settings. 

In summary, the results confirm that \FuncCall{SimAdjust}{} and \FuncCall{DistAdjust}{} complement each other, improving attack effectiveness while maintaining stealthiness when used together.

\section{Additional Results with Label-flipping Attack}
In addition to the untargeted model poisoning attacks discussed in the main text, we evaluate our method using a data-poisoning-based targeted attack: the label-flipping attack. Label-flipping is a straightforward yet effective targeted attack in federated learning (FL) and is also among the least stealthy data-poisoning-based attacks. To make the evaluation more challenging, we configured the attacker to flip all the labels of the training data on malicious clients, making the label-flipping attack even less stealthy. The results are shown in Table~\ref{Table:label flip}, demonstrating that our method enhances the label-flipping attack to bypass six additional defenses compared to its original version. This illustrates the compatibility of our method for data-poisoning-based and targeted attacks.

\section{Additional Results with More Complex Models}
To further demonstrate the effectiveness of our method on more complex model architectures, we test our method using VGG-11 net on the IID CIFAR-10 dataset, shown in Table~\ref{Table:vgg}. The results demonstrate that our method consistently enhances the performance of Sign-flipping Attack, Trim Attack, and Krum Attack, outperforming their original versions in 41 out of 54 cases. These findings illustrate the effectiveness of our method with more complex model architectures.

\section{Additional Results with Larger FL Systems}
\label{appendix: scalability}
To further demonstrate the effectiveness of our method in larger FL systems, we extend our experiments on the Fashion-MNSIT dataset with 100 clients, shown in Table~\ref{Table:100-trim}. The results show a similar trend as observed in the 50-client system. Our method enables baseline attacks to successfully bypass four additional defenses, causing over 50\% additional error rates in the global model. These findings further validate the effectiveness and generality of our approach in larger systems, when a single malicious client has fewer data samples.

\section{Comparison with Existing Attack Enhancement Method}
Considering several prior studies~\citep{bagdasaryan2020backdoor,bhagoji2019analyzing,zhang2022neurotoxin} enhancing backdoor attacks, we also adapt one recent one - Neurotoxin~\citep{zhang2022neurotoxin} - to our untargeted attacks evaluation setting. Table~\ref{Table:neurotoxin} illustrates the results on the IID Fashion-MNIST dataset within a 50-client FL system using Trim Attack. The results demonstrate that our method outperforms Neurotoxin in 24 out of 28 cases. This highlights that directly transferring existing methods designed for backdoor attacks may not yield consistent effectiveness when applied to untargeted attack scenarios. The results further validate the robustness of our approach.

\begin{table*}[t]
	\centering
    \small
        \setlength{\tabcolsep}{1.8pt}
	    \begin{tabular}{cccccccc|ccccccc}
		\toprule
        \textbf{Data Distribution} & \multicolumn{7}{c}{\textbf{IID}} & \multicolumn{7}{c}{\textbf{Non-IID}}\\
        \otoprule
		\textbf{Attack} & \textbf{FedAvg} & \textbf{FLTrust} & \textbf{MKrum} & \textbf{Bulyan} & \textbf{Median} & \textbf{Trim} & \textbf{FLD} & \textbf{FedAvg} & \textbf{FLTrust} & \textbf{MKrum} & \textbf{Bulyan} & \textbf{Median} & \textbf{Trim} & \textbf{FLD}\\
		\otoprule
		No Attack & 0.109 & 0.107 & 0.105 & 0.105 & 0.123 & 0.106 & 0.115 & 0.113 & 0.115 & 0.115 & 0.112 & 0.142 & 0.115 & 0.122 \\
        \midrule
		Sign-Flipping & \textbf{0.935} & \textbf{0.123} & 0.098 & 0.101 & 0.106 & 0.101 & 0.103 & \textbf{0.903} & 0.114 & 0.103 & 0.107 & 0.124 & 0.111 & 0.116 \\
		\cmidrule{2-15}
        \rowcolor{BlueGray2}
        \cellcolor{White} + Poison Pill & 0.153 & 0.104 & \textbf{0.216} & \textbf{0.195} & \textbf{0.251} & \textbf{0.146} & \textbf{0.584} & 0.155 & \textbf{0.115} & \textbf{0.177} & \textbf{0.384} & \textbf{0.85}1 & \textbf{0.284} & \textbf{0.549} \\
        \midrule
        Trim Attack & 0.102 & \textbf{0.112} & 0.101 & 0.106 & 0.113 & 0.103 & 0.095 & 0.118 & \textbf{0.113} & 0.118 & 0.113 & 0.136 & 0.114 & 0.108 \\
		\cmidrule{2-15}
        \rowcolor{BlueGray2}
        \cellcolor{White} + Poison Pill & \textbf{0.206} & 0.102 & \textbf{0.285} & \textbf{0.163} & \textbf{0.314} & \textbf{0.109} & \textbf{0.300} & \textbf{0.329} & 0.109 & \textbf{0.284} & \textbf{0.288} & \textbf{0.362} & \textbf{0.165} & \textbf{0.344} \\
		\bottomrule
	    \end{tabular}
    \caption{Performance when the number of malicious clients is gradually decreasing.}
	\label{Table:gradual decrease}
\end{table*}
\section{Results with Decreasing Number of Malicious Clients}
\label{appendix:decreasing mal}
To further demonstrate the effectiveness of our method in a more practical setting, we evaluate its performance as the number of malicious clients in the FL system gradually decreases. Specifically, we used the Fashion-MNIST dataset within a 50-client FL system. Initially, 20\% of clients are malicious, and for every T/4 rounds (where T is the total number of FL communication rounds), the proportion of malicious clients reduces by 5\%. Here is a detailed breakdown of this setup:
\begin{itemize}
    \item $0\rightarrow \frac{T}{4}$: 20\% clients in the FL system are malicious.
    \item $\frac{T}{4}\rightarrow \frac{T}{2}$: 15\% clients in the FL system are malicious.
    \item $\frac{T}{2}\rightarrow \frac{3T}{4}$: 10\% clients in the FL system are malicious.
    \item $\frac{3T}{4}\rightarrow T$: 5\% clients in the FL system are malicious.
\end{itemize}
The results are presented in Table~\ref{Table:gradual decrease}, demonstrating that our method significantly enhances the error rates achieved by the original Trim Attack and Sign-flipping attack in 23 out of 28 cases, with an average error rate increase of over 50\%. These findings illustrate the robustness and effectiveness of our method in a more practical scenario.